\documentclass{article}

\usepackage{microtype}
\usepackage{graphicx}
\usepackage{subcaption}
\usepackage{booktabs} 
\usepackage{hyperref}
\usepackage{multirow}
\usepackage{booktabs} 
\usepackage{enumitem}
\usepackage{float}
\usepackage{amsmath}
\usepackage{amssymb}
\usepackage{mathtools}
\usepackage{amsthm}
\usepackage{wrapfig}
\usepackage{xspace}
\usepackage{setspace}
\usepackage{threeparttable}
\usepackage[ruled,vlined]{algorithm2e}

\newcommand{\method}{TraceNAS}

\usepackage[preprint]{icml2026}

\usepackage[compact]{titlesec}
\titlespacing{\section}{0pt}{0.35ex}{0.15ex}
\titlespacing{\subsection}{0pt}{0.2ex}{0.13ex}
\titlespacing{\paragraph}{0pt}{-1ex}{1em}

\usepackage[capitalize,noabbrev]{cleveref}

\usepackage{etoc}

\theoremstyle{plain}

\theoremstyle{definition}

\theoremstyle{remark}

\usepackage[textsize=tiny]{todonotes}
\usepackage{etoc}

\icmltitlerunning{TraceNAS: Zero-shot LLM Pruning via Gradient Trace Correlation}

\begin{document}

\twocolumn[
    \icmltitle{TraceNAS: Zero-shot LLM Pruning via Gradient Trace Correlation}
    
    
    
    
    \icmlsetsymbol{equal}{*}
    \icmlsetsymbol{current}{$\ddagger$} 
    
    \begin{icmlauthorlist}
    \icmlauthor{Prajna G. Malettira}{purdue}
    \icmlauthor{Manish Nagaraj}{purdue}
    \icmlauthor{Arjun Roy}{purdue}
    \icmlauthor{Shubham Negi}{purdue,google,current}
    \icmlauthor{Kaushik Roy}{purdue}
    \end{icmlauthorlist}
    
    \icmlaffiliation{purdue}{Purdue University}
    \icmlaffiliation{google}{Google \textsuperscript{$\ddagger$}Work done while at Purdue University}
    \icmlcorrespondingauthor{Prajna G. Malettira}{pmaletti@purdue.edu}
    
    
    \vskip 0.3in
]



\printAffiliationsAndNotice{}  


\begin{abstract}
    Structured pruning is essential for efficient deployment of Large Language Models (LLMs). The varying sensitivity of LLM sub-blocks to pruning necessitates the identification of optimal non-uniformly pruned models. Existing methods evaluate the importance of layers, attention heads, or weight channels in isolation. Such localized focus ignores the complex global structural dependencies that exist across the model. Training-aware structured pruning addresses global dependencies, but its computational cost can be just as expensive as post-pruning training. To alleviate the computational burden of training-aware pruning and capture global structural dependencies, we propose \method, a training-free Neural Architecture Search (NAS) framework that jointly explores structured pruning of LLM depth and width. \method{} identifies pruned models that maintain a high degree of loss landscape alignment with the pretrained model using a scale-invariant zero-shot proxy, effectively selecting models that exhibit maximal performance potential during post-pruning training. \method{} is highly efficient, enabling high-fidelity discovery of pruned models on a single GPU in 8.5 hours, yielding a 10$\times$ reduction in GPU-hours compared to training-aware methods. Evaluations on the Llama and Qwen families demonstrate that \method{} is competitive with training-aware baselines across commonsense and reasoning benchmarks.
\end{abstract}

{\setlength{\abovedisplayskip}{4pt}
\setlength{\belowdisplayskip}{4pt}
\begin{figure}[t]
    \centering
    \includegraphics[width=0.9\columnwidth]{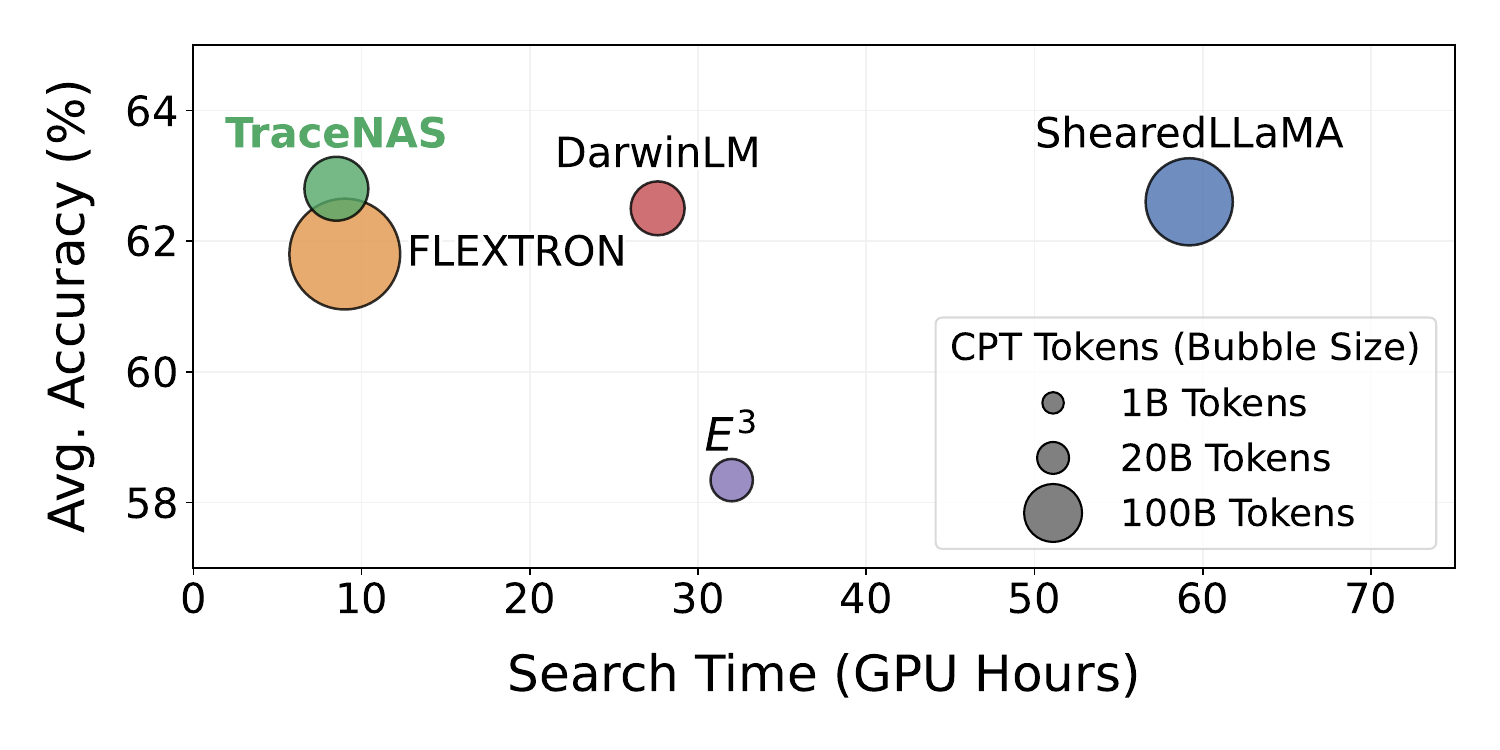}
    \vspace{-2ex}
    \caption{\textbf{Search Efficiency.} \method{} identifies optimal non-uniform architectures in 8.5 GPU hours\footnotemark[3], achieving competitive accuracy with 10$\times$ less search overhead than training-aware baselines. The area of the bubble is proportional to the total tokens used for recovery training, highlighting that \method{} identified architectures have high recovery potential.}
    \vspace{-4ex}
    \label{fig:search_efficiency}
\end{figure}}
\section{Introduction} \label{sec:intro}
Recent breakthroughs in Large Language Models (LLMs)~\cite{anthropic2023introducing, openai2023gpt} have delivered unprecedented capabilities but require massive scale in parameters and computation. Foundational scaling laws~\cite{kaplan2020scaling} have established that smaller LMs are less sample-efficient, and thus training them from scratch is a resource-intensive bottleneck. Model compression enables the realization of efficient models that preserve the extensive knowledge of the pretrained base. Techniques such as quantization~\cite{frantar2022gptq, liu2024llm}, distillation~\cite{sarah2024llama,li2022shadow} and pruning~\cite{xia2023sheared, sun2023simple} have become essential, allowing for the accessible deployment of high-performance architectures. Among these, structured pruning~\cite{ma2023llm} aims to remove architectural units like attention heads, weight columns, or transformer blocks to achieve immediate hardware speedup. This process can be naturally formulated as a Neural Architecture Search (NAS)~\cite{sieberlingevopress,tao-etal-2023-structured,tang2025darwinlm} problem, where the pretrained LLM is conceptualized as a super-network~\cite{yu2020bignas,cai2019once} that serves as the search space.

\footnotetext[3]{\method{}, ShearedLlama and Flextron report search time on a NVIDIA A100, DarwinLM on L40S and E$^3$ reports NPU hours.}
Structured pruning is inherently more disruptive than the localized masking used in unstructured pruning~\cite{sun2023simple, an2024fluctuation, frantar2023sparsegpt}. While unstructured methods leave the model's architecture intact, structured pruning physically alters it and risks severing critical pathways in the representational flow. The primary bottleneck in adapting unstructured methods to structured pruning is the lack of a comparable global metric across the model's layers~\cite{ma2023llm}. Local importance heuristics such as activation-weighted magnitudes~\cite{sun2023simple,an2024fluctuation} or Hessian-based reconstructions~\cite{frantar2023sparsegpt} identify redundancy within isolated layers and are often incomparable across model depth due to varying magnitude scales~\cite{ma2023llm}. We address this limitation by utilizing gradients~\cite{das2023beyond,yang2025wanda++} as global signals that naturally capture the complex inter-dependencies across the model.

Previous search methods for structured pruning employ mask learning~\cite{xia2023sheared, yuan20253} or training-aware search~\cite{tang2025darwinlm, pmlr-v267-bercovich25a}. While these training-aware approaches define the state-of-the-art, they demand substantial computational resources. This overhead stems from the iterative gradient updates required for \emph{training-aware} model discovery and also necessitates a continuous stream of unseen tokens to prevent pruned models from overfitting to calibration data. This cost can make the search for an efficient model just as demanding as post-pruning recovery, motivating the need for \emph{training-free NAS}~\cite{wu2021training,tran2021feature,ingolfsson2022reducing} using zero-shot performance proxies. However, existing proxies are primarily designed to identify trainable, randomly initialized architectures by focusing on heuristics such as gradient stability~\cite{abdelfattah2021zero} or loss landscape smoothness~\cite{li2023zico}. These metrics are insufficient for LLM pruning as they prioritize trainability, the ability of a network to learn from scratch. In contrast, the performance of pruned LLMs depends on inheritance, the ability of the pruned model to recover performance by preserving the pretrained loss landscape~\cite{frankle2018lottery,frankle2020linear,chen2020lottery}.

In this paper, we introduce \method, a training-free NAS framework for joint depth and width pruning of LLMs. Our approach leverages the observation that a pretrained model resides in broad, stable regions of the loss landscape~\cite{frankle2020linear,li2018visualizing}. We propose a zero-shot proxy that utilizes gradient trace to capture the impact of structural pruning on model sensitivity. Specifically, the proxy measures the sparsity-weighted aggregate over the Pearson Correlation coefficients between the gradient traces of the pruned and pretrained base models. Since structured pruning alters activation and gradient magnitudes, the use of Pearson Correlation ensures scale invariance, thus capturing the alignment of gradient traces on the pretrained loss landscape independently of scale shifts.

\begin{figure*}[t]
    \centering
    \includegraphics[width=0.8\textwidth]{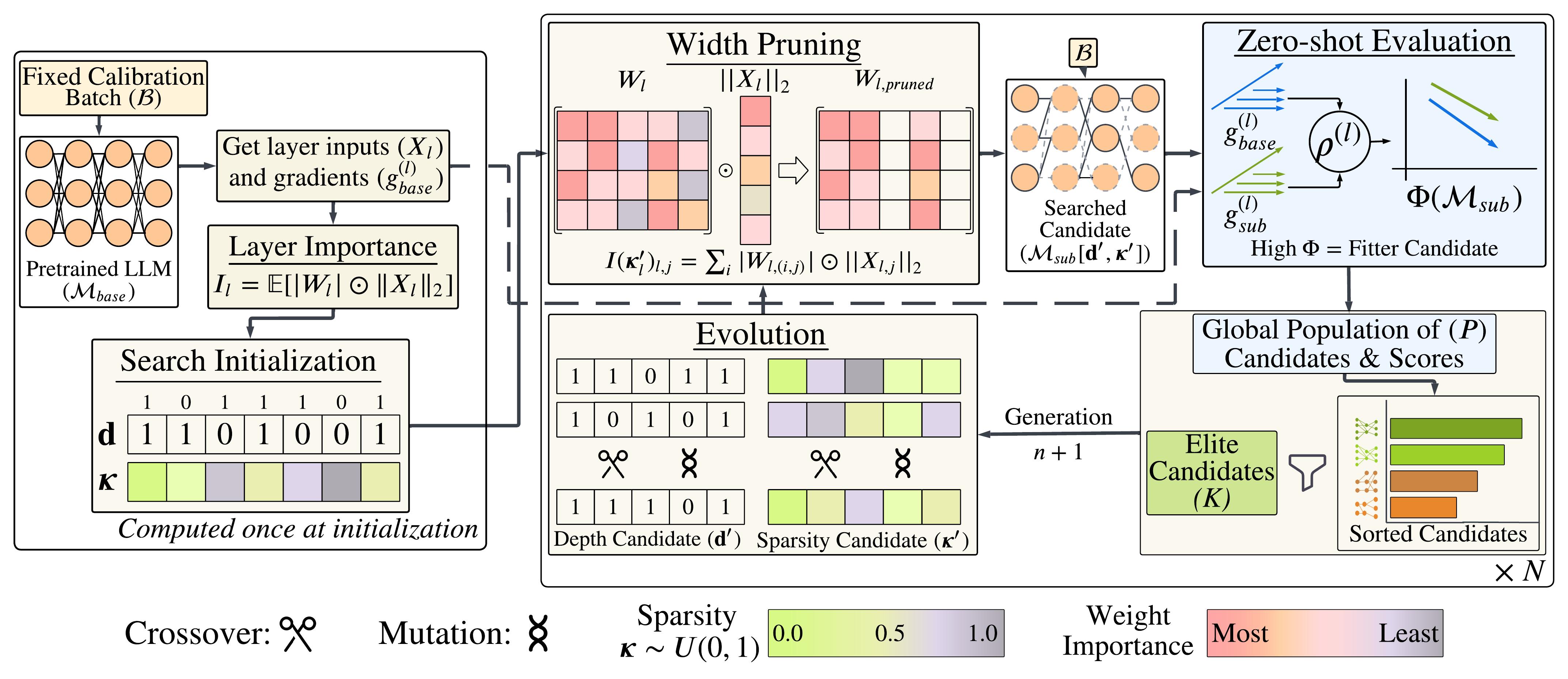}    
    \caption{\textbf{Visualization of the \method{} search framework.} \method{} uses a gradient-based, training-free proxy to guide structural pruning. Following a one-time initialization of base gradient traces ($g_{\text{base}}$) and importance scores ($I_l$), a population of depth ($\mathbf{d}$) and width ($\boldsymbol{\kappa}$) candidates undergoes iterative evolution via crossover and mutation. Each width configuration is realized using an $O(d^2)$ activation-weighted heuristic. Subsequently, candidates $\mathcal{M}_{sub}$ are ranked by the zero-shot proxy $\Phi$, which measures the gradient trace alignment between the active layers of $\mathcal{M}_{sub}$ relative to $\mathcal{M}_{base}$.}
    \vspace{-3ex}
    \label{fig:methodology_overview}
\end{figure*}

To ensure the search remains tractable across varying model scales, we compute these gradients within a low-rank~\cite{hu2022lora,zhao2024galore} subspace. By leveraging the intrinsic low-dimensionality~\cite{aghajanyan-etal-2021-intrinsic} of LLMs, \method{} captures the principal gradient components of the model while reducing memory overhead by orders of magnitude. Unlike training-aware methods which require significant memory overhead to maintain optimizer states and activations for backpropagation across a large set of calibration tokens, our training-free approach only requires the gradients of each candidate computed on a minimal calibration set. This reduction in memory consumption enables high-fidelity model discovery (Fig.~\ref{fig:search_efficiency}), bypassing the compute intensive requirements typical of training-aware NAS. The primary contributions of this work are: 
\begin{enumerate}[itemsep=1pt, parsep=0pt, topsep=1pt] 
    \item \textbf{Scalable Training-free NAS Framework:} We introduce \method{}, a unified pruning framework (Fig.~\ref{fig:methodology_overview}) for the joint optimization of LLM depth and width. Our approach enables the zero-shot discovery of non-uniform pruned models, facilitating efficient compression across model scales without search-time training.
    \item \textbf{Zero-shot Proxy for Functional Inheritance:} We introduce a scale-invariant, gradient-based zero-shot proxy that evaluates gradient alignment between the pruned and pretrained models. Our proxy achieves superior Spearman Rho~\cite{spearman1961proof} ($\rho=0.94$) and Kendall Tau~\cite{kendall1938new} ($\tau=0.82$) correlations with downstream accuracy, enabling accurate zero-shot discovery.
    \item \textbf{Efficient Architecture Discovery:} We show that \method{} achieves a 10$\times$ reduction in both GPU-hours and total calibration data compared to training-aware methods (Fig.~\ref{fig:search_efficiency}). This drastically lowers the overhead for large-scale model compression by ensuring search costs are significantly lower than recovery training.
\end{enumerate}
\section{Related Work}
\label{sec:related_work}
\subsection{Language Model (LM) Pruning}
\label{subsec:related_work:lm_pruning}
LLM pruning methods vary by granularity: unstructured weight sparsification~\cite{frantar2023sparsegpt} or structured removal~\cite{ma2023llm,kim2024shortened} of transformer blocks, attention heads or weight channels. These methods are further categorized by their computational complexity relative to the model's hidden dimension $d$. Early $O(d)$ magnitude pruning~\cite{han2015deep} methods evaluate weights in isolation, ignoring the outlier features characteristic of LLMs~\cite{kovaleva2021bert,luo2021positional,yin2023outlier}. To preserve these outliers, $O(d^2)$ metrics like Wanda~\cite{sun2023simple} and FLAP~\cite{an2024fluctuation} utilize weight-activation products. Furthermore, $O(d^3)$ optimization-based methods like SparseGPT~\cite{frantar2023sparsegpt} and LLM-Pruner~\cite{ma2023llm} utilize second-order inverse Hessians~\cite{hassibi1993optimal} to minimize layer-wise reconstruction error. However, these local heuristics prioritize layer properties and fail to account for how pruning-induced errors propagate through the model.

\subsection{Influence-based Importance}
Beyond these heuristics, recent research has explored influence functions (IF)~\cite{koh2017understanding, kwon2023datainf} to better capture how model perturbations affect sensitivity. For instance, LayerIF~\cite{askari2025layerif} uses IF to estimate layer importance; however, it remains locally heuristic as it perturbs layers individually to measure validation loss sensitivity. This inherently fails to capture the joint impact of multi-layer and width pruning on the model's representation. Furthermore, the $O(d^3)$ complexity of standard IF is prohibitive for LLMs, requiring inverse Hessian computations for all model parameters. In contrast, \method{} adopts an efficient first-order gradient-tracing approach inspired by TracIn~\cite{pruthi2020estimating}  to evaluate the global impact of pruning. By operating with $O(d^2)$ complexity, \method{} identifies non-uniform pruned architectures that maintain functional alignment with the pretrained base model, bypassing the computational bottlenecks and local limitations of LayerIF. We provide a detailed comparison between IF, IF-based pruning and \method{} in~\ref{appendix:compare_influence_functions}.

\subsection{Training-Aware LM Pruning} 
\label{subsec:related_work:training_aware_pruning}
Recent structured pruning methods, such as ShearedLlama~\cite{xia2023sheared} and $\text{E}^3$~\cite{yuan2025e3prunerefficienteconomicaleffective}, use mask learning via $L_0$ regularization and differential mask optimization. While effective, these methods target uniform sparsity across layers, failing to account for the heterogeneous importance distribution across a model's width~\cite{tao-etal-2023-structured}. This has prompted a shift toward discovering non-uniform architectures via evolution search. Methods like SIMPLE~\cite{tao-etal-2023-structured} and EvoPress~\cite{sieberlingevopress} employ evolutionary optimization for heterogeneous discovery.  On the other hand, DarwinLM~\cite{tang2025darwinlm} uses a curriculum-style training-aware search, performing lightweight fine-tuning for every searched candidate. Similarly, PUZZLE~\cite{pmlr-v267-bercovich25a} utilizes distillation-based NAS to identify optimal configurations by minimizing block-wise KL divergence. However, the training-aware search inherent to these methods mean their model discovery can be just as computationally intensive as the post-pruning recovery training.

\subsection{Zero-shot NAS for LM Pruning} 
As models scale, these training-intensive searches become intractable, necessitating training-free solutions for NAS. Existing proxies are formulated to identify trainable, randomly initialized architectures by focusing on heuristics such as gradient stability~\cite{abdelfattah2021zero}, model expressivity~\cite{mellor2021neural,jiang2023meco}, or loss landscape smoothness~\cite{li2023zico}. These proxies are insufficient for evaluating pruned models as they do not account for how well a model inherits pretrained knowledge. To address this, recent works LPZero~\cite{dong2024lpzero} and Pruner-Zero~\cite{dong2024pruner} formulate zero-shot pruned model discovery as a two-fold problem: first searching for unique symbolic pruning metrics for the target pretrained model, and subsequently applying these metrics to discover the optimal structured~\cite{dong2024lpzero} or unstructured~\cite{dong2024pruner} sub-network. While modular, the dual-stage process adds overhead to the search pipeline. In contrast, \method{} introduces a unified training-free fitness proxy that directly evaluates a compressed model's functional alignment with its pretrained base, streamlining the search for pruned models that have high performance potential and fast convergence during post-pruning recovery.
\section{Methodology}
\subsection{Problem Definition} 
\label{subsec:problem_def}
We formulate pruned model discovery as a constrained discrete search over a pretrained super-network, $\mathcal{M}_{base}$~\cite{cai2019once,yu2020bignas}. We seek an optimal $\hat{\mathcal{M}}_{sub} \in \mathcal{G}$ that maximizes a training-free proxy $\Phi(\cdot,\cdot)$, that measures functional inheritance; the ability of $\mathcal{M}_{sub}$ to recover performance by preserving the gradient direction of $\mathcal{M}_{base}$ on it's loss landscape:
{\setlength{\abovedisplayskip}{4pt}
\setlength{\belowdisplayskip}{4pt}
\begin{equation*}
    \hat{\mathcal{M}}_{sub} =  \mathop{\arg\max}\limits_{\mathcal{M} \in \mathcal{G}} \; \Phi(\mathcal{M}_{sub}, \mathcal{M}_{base}) \; \text{s.t.} \; \mathcal{P}(\mathcal{M}_{sub}) \leq C,
\end{equation*}}

where $\mathcal{G}$ is the search space, $\mathcal{P}(\cdot)$ denotes the parameter count, and $C$ is the target constraint. By maximizing $\Phi$, we identify models with maximal potential for performance recovery during post-pruning \textit{continued pretraining} (CPT) without the overhead of search-time training.

\subsection{Search Space Encoding}
\label{subsec:search_space_encoding}
The search space $\mathcal{G}$ is defined as a joint distribution over the model's depth and sub-block-wise width, enabling the discovery of non-uniformly pruned architectures from $\mathcal{M}_{base}$. By operating at sub-block granularity, the \method{} search framework treats attention and MLP modules as independent units, thereby capturing diverse structural sparsity.

\paragraph{Architecture Encoding} A candidate $\mathcal{M}_{sub}$ is parameterized by a tuple $(\mathbf{d}, \boldsymbol{\kappa})$. The depth configuration $\mathbf{d} \in \{0,1\}^L$ is a discrete mask over the $L$ blocks of $\mathcal{M}_{base}$. Setting $d_l=0$ deactivates the $l$-th block while preserving the residual stream, allowing the search to explore varying model depths. We enforce a minimum depth constraint, $\sum d_l \geq L_{min}$, to ensure sufficient computational capacity for sequential reasoning. Simultaneously, the width configuration $\boldsymbol{\kappa}=\{ \kappa_1, \dots, \kappa_L \}$ where $\kappa_l = (r_{attn}^{(l)}, r_{mlp}^{(l)}) \in (0, 1]$ defines the parameter retention ratio for each attention and MLP sub-block. By decoupling the sparsity ratios, our framework identifies heterogeneous architectures that distinctly prioritize retention in sensitive sub-modules while respecting the global budget $C$.

\subsection{In-Place Architectural Realization} 
\label{subsec:realization}
We realize candidate sub-networks $\mathcal{M}_{sub}$ by mapping sparsity ratios $\boldsymbol{\kappa}$ to binary masks and implementing them via an efficient in-place masking strategy.

\paragraph{Width Mask Generation} To realize a layer's width configuration $\kappa_l$ into a weight mask, we use the activation-weighted product: $I(\kappa_l)_{l,j} = \sum_i |W_{l,(ij)}| \odot \|X_{l,j}\|_2$~\cite{sun2023simple} as a saliency metric. This approach explicitly accounts for the outlier features characteristic of LLMs~\cite{kovaleva2021bert,luo2021positional, yin2023outlier} by scaling weight magnitudes by the $L_2$-norm of their corresponding input activations. By identifying channels that are functionally critical to the model's representational flow, $I(\kappa_l)_{l,j}$ ensures that mask realizations are tailored to preserve the underlying representation of $\mathcal{M}_{base}$. 

Adopting this $O(d^2)$ heuristic avoids the $O(d^3)$ bottleneck of second-order reconstruction methods~\cite{frantar2023sparsegpt, ma2023llm,tang2025darwinlm}, enabling the high-frequency scoring and realization of hundreds of distinct candidates during the evolutionary search. Crucially, while $I(\kappa_l)_{l,j}$ provides a local channel-selection heuristic, the global distribution of the sparsity ratios $\kappa_l$ is governed by the \method{} proxy $\Phi$ (Sec.~\ref{subsec:tracenas_score}), which accounts for the inter-layer dependencies and impact of pruning that local heuristics ignore.

\paragraph{In-Place Masking} To score pruned candidates, we use an in-place masking strategy that operates directly on the weight matrices of $\mathcal{M}_{base}$. This eliminates the need for  instantiating the population of candidate models during each search iteration. For depth exploration, $\mathcal{M}_{base}$ is modified according to $d_l \in \mathbf{d}$; if $d_l = 0$, the entire block is bypassed and activations are routed through the residual connection, ensuring functional continuity~\cite{veit2016residual,gromov2024unreasonable}. For active blocks $(d_l = 1)$ the binary masks are applied via a temporary modify-then-restore pointer: $W'_{base} = W_{base} \odot \text{Mask}(\kappa_l)$. This allows the search framework to execute forward and backward passes over a calibration set to capture the cumulative effects of pruning without the memory overhead of instantiating $\mathcal{M}_{sub}$. Once the gradient trace is recorded for scoring, the original weights are restored, preserving $\mathcal{M}_{base}$ for subsequent search iterations.

\subsection{\method: Evaluating Functional Alignment}
\label{subsec:tracenas_score}
To evaluate the functional integrity of realized candidates $\mathcal{M}_{sub}$, we propose the proxy $\Phi$. Structured pruning induces a significant representational shift that manifests as an immediate drop in performance~\cite{tran2022pruning}. This initial shock obscures the distinction between fundamental structural collapse and architectures that preserve the essential representational pathways required for effective post-pruning recovery.

Furthermore, as discussed in Sec.~\ref{subsec:related_work:lm_pruning}, importance-based metrics~\cite{sun2023simple,an2024fluctuation} prioritize local weight properties while overlooking the structural inter-dependencies captured by gradients~\cite{das2023beyond}. To address this, we define $\Phi$ as a zero-shot, gradient-based proxy that quantifies functional inheritance. $\Phi$ measures the directional alignment between the gradient trace of $\mathcal{M}_{sub}$ and $\mathcal{M}_{base}$. This directional alignment allows $\Phi$ to identify models that are favorably positioned within the pretrained convergence basin, thereby enhancing their potential for functional recovery.

\paragraph{Establishing the Functional Anchor} We define the optimal functional state as the gradient trace $g_{base}$ of the pretrained model $\mathcal{M}_{base}$. Using a calibration set $\mathcal{B}$, $g_{base}$ provides the directional anchor within the optimization landscape that sub-networks $\mathcal{M}_{sub}$ must align with to ensure functional inheritance. The gradient traces for the base and candidate models are calculated as 
{\setlength{\abovedisplayskip}{4pt}
\setlength{\belowdisplayskip}{4pt}$$g = \mathbb{E}_{b \in \mathcal{B}} [\nabla_{\theta} \mathcal{L}(\mathcal{M}(b; \theta))]$$}
where $\theta$ denotes the trainable model parameters. 

\paragraph{Low-Rank Gradient Manifold} To ensure computationally tractability across model sizes, we compute gradient traces $g$ within a low-rank subspace. Since storing full-rank LLM gradients is memory-prohibitive, we attach Low-Rank Adapters (LoRA)~\cite{hu2022lora} to all linear projections. This approach leverages the observation that the functional knowledge of LLMs resides within a low-dimensional manifold~\cite{aghajanyan-etal-2021-intrinsic, zhao2024galore}, allowing us to capture global dependencies while significantly reducing the memory footprint of gradient storage.

\paragraph{Measuring Functional Alignment} To quantify how effectively $\mathcal{M}_{sub}$ inherits the functional state of $\mathcal{M}_{base}$, we calculate the Pearson Correlation Coefficient $\rho^{(l)}$ per layer $l \in L$. For each transformer-block $l$, $\rho^{(l)}$ measures the alignment between the low-rank gradient traces of the pruned attention and MLP sub-blocks:
{\setlength{\abovedisplayskip}{4pt}
\setlength{\belowdisplayskip}{4pt}
\begin{equation} \rho^{(l)} = \frac{1}{N_l} \left\langle \left( \frac{g_{sub}^{(l)} - \mu_{g_{sub}^{(l)}}}{\sigma_{g_{sub}^{(l)}}} \right), \left( \frac{g_{base}^{(l)} - \mu_{g_{base}^{(l)}}}{\sigma_{g_{base}^{(l)}}} \right) \right\rangle \end{equation}}
where $N_l$ is the dimensionality of the low-rank subspace, and $\mu, \; \sigma$ denote the mean and standard deviation of the gradient elements. By standardizing the traces, we decouple the directional alignment from magnitude shifts inherently induced by structural pruning.

\paragraph{Sparsity-Weighted Aggregation} We formulate the final proxy $\Phi$ by aggregating sub-block correlations using a sparsity-weighted summation. While gradient standardization ensures local scale-invariance, a uniform average of $\rho^{(l)}$ across layers fails to account for the heterogeneous representation capacity of $\mathcal{M}_{sub}$. We therefore weight each correlation coefficient by its corresponding retention ratio $r^{(l)} \in \kappa_l$:
{\setlength{\abovedisplayskip}{4pt}
\setlength{\belowdisplayskip}{4pt}
\begin{equation}
\Phi(\mathcal{M}_{sub},\mathcal{M}_{base}) = \sum_{l \in \text{Attn}} r_{attn}^{(l)} \cdot \rho^{(l)} + \sum_{l \in \text{MLP}} r_{mlp}^{(l)} \cdot \rho^{(l)}
\end{equation}}

This formulation anchors the global score in the high-capacity regions of $\mathcal{M}_{sub}$, prioritizing the sub-blocks that serve as the primary repositories of functional inheritance. By weighting $\rho^{(l)}$ by parameter density, we prevent $\Phi$ from being skewed by the high-variance noise typical of aggressively pruned sub-blocks.

\subsection{Selection \& Evolutionary Mechanics} 
\label{subsec:search_mechanics}
To navigate the discrete search space $\mathcal{G}$, we employ an evolutionary strategy driven by the proxy $\Phi$. The initial search population $P_0$ is generated using a global importance prior; detailed initialization procedures and a search convergence analysis are provided in Appendix~\ref{appendix:search_space_init}.

\paragraph{Search Space Evolution} We employ a hybrid evolution strategy for the joint $(\mathbf{d}, \boldsymbol{\kappa})$ search space. Depth evolution explores discrete block configurations via contiguous-range crossover and bit-flip mutations. Simultaneously, width evolution explores continuous sparsity ratios using interpolation crossover and multiplicative Gaussian jitter-based mutation. This joint evolution allows \method{} to explore the architectural trade-offs between depth and width while adhering to the constraint $C$.

\paragraph{Search Granularity} The activations used for mask generation are obtained at the attention output ($W_o$) and MLP down ($W_{down}$) projection layers. These projections consolidate sub-block representations, and their input activations serve as proxies for the importance of preceding upstream weights. Once generated, these masks are applied to the corresponding projections: $W_q, W_k, W_v$ for attention and $W_{up}, W_{gate}$ for MLP. For Grouped-Query Attention (GQA)~\cite{ainslie2023gqa} models, we do not prune $W_k$ and $W_v$ to ensure KV-cache compatibility. Furthermore, pruned MLP hidden dimensions are rounded to multiples of $m=32$ to maintain efficient tensor operations.

\subsection{Interpreting Functional Inheritance through Gradient Trace Alignment}
The effectiveness of $\Phi$ as a proxy for functional inheritance is driven by three key observations:
\begin{itemize}[itemsep=1pt, parsep=0pt, topsep=1pt] 
    \item \textbf{Manifold Anchoring:} High-performing pretrained models reside within flat, stable convergence basins~\cite{frankle2020linear,li2018visualizing}. A high alignment score $\Phi$ implies that the first-order Taylor expansion of the sub-network’s loss surface remains congruent with that of the base model. By preserving the gradient's directional signature, $\mathcal{M}_{sub}$ remains anchored within the original functional manifold. This ensures that post-pruning recovery initiates from a region of high directional certainty, enabling a stable loss trajectory rather than necessitating a costly search for a new local minimum.
    \item \textbf{Global Sensitivity via Chain-Rule:} Unlike metrics that evaluate layers in isolation~\cite{sun2023simple,askari2025layerif}, $g_{sub}$ is computed via backpropagation through the masked computational graph. This captures inter-layer disruptions: if an upstream block bottlenecks the representational flow, subsequent layer gradient traces will de-correlate from the base model optimization manifold. This causes a sharp drop in $\Phi$, allowing the proxy to detect structural incoherence and broken residual streams that local heuristics are inherently blind to.
    \item \textbf{Intrinsic Dimensionality Invariance:} Empirical evidence shows that gradient alignment is largely rank-invariant (Fig.~\ref{fig:stability_triple}c, Appendix~\ref{appendix:corr_amoaballm_stability}). This indicates that $\Phi$ can capture a pruned model's functional capability within a low-dimensional~\cite{aghajanyan-etal-2021-intrinsic} manifold. This robustness ensures that the relative performance ranking of candidates remains stable across gradient ranks~\cite{zhao2024galore}, justifying the use of low-rank subspaces as a reliable and scalable surrogate for full-rank analysis.
\end{itemize}
\section{Experiments}
\label{sec:experiments}
\subsection{Experimental Setup}
\paragraph{Models and Datasets:} We evaluate \method{} across multiple scales using Llama-2-7B~\cite{touvron2023llama}, Llama-3.1-8B~\cite{dubey2024llama} and Qwen-2.5-14B, covering both Multi-Head Attention (MHA)~\cite{vaswani2017attention} and Grouped-Query Attention (GQA)~\cite{ainslie2023gqa} architectures. We use the FineWeb-Edu~\cite{penedo2024fineweb} 100BT dataset for both evolutionary search and post-pruning CPT. For the search, we use a fixed calibration set of 65,536 tokens (16 sequences) per pruned candidate to evaluate a total population of 1500 candidates. This is followed by 20B tokens of post-pruning CPT. Results on additional models and CPT token budgets is provided in Appendix~\ref{appendix:tracenas_model_scales} and~\ref{appendix:token_scales}.

\paragraph{Implementation Details} The evolutionary search is performed with a population of 30 candidates over 50 search iterations. On a single NVIDIA H200 GPU, the search concludes in $\sim$2 hours, a significant acceleration over the 8.5-hour A100 baseline reported in Fig.~\ref{fig:search_efficiency}. The highest scored sub-networks then undergo CPT on a cluster of six H200 GPUs ($\sim$16 hours) using a global batch size of 1024 and peak learning rate of $1e^{-4}$. We use a Warmup-Stable-Decay (WSD)~\cite{hu2024minicpm} scheduler, with context lengths set to 4096 for Llama-2 and Qwen-2.5, and 8192 for Llama-3.1. Detailed hyperparameters are available in Appendix~\ref{appendix:hyperparams}.

\paragraph{Baselines:} We benchmark \method{} against leading structured and layer-pruning methods. This includes, mask-learning methods, ShearedLLaMA~\cite{xia2023sheared} and $E^3$-Pruner~\cite{yuan20253}, which employ Lagrangian optimization. For training-aware pruning, we compare against DarwinLM~\cite{tang2025darwinlm}. We further evaluate against Minitron~\cite{sreenivas2024llm}, Flextron~\cite{cai2024flextron}, LoRAP~\cite{li2024lorap} and uniformly pruned models found via \method{} to isolate the benefits of non-uniform pruning. Comparisons with unstructured pruning methods like Wanda~\cite{sun2023simple}, FLAP~\cite{an2024fluctuation} and PrunerZero~\cite{dong2024pruner} are included in Appendix~\ref{appendix:compare_unstructured}.

\paragraph{Evaluation Benchmarks:} Models are evaluated using the \texttt{lm-evaluation-harness}~\cite{eval-harness} across 0-shot commonsense reasoning benchmarks: ARC-easy~\cite{clark2018think}, LogiQA~\cite{liu2020logiqa}, PIQA~\cite{bisk2020piqa}, SciQ~\cite{welbl2017crowdsourcing} and BoolQ~\cite{clark2019boolq}. We also report 5-shot performance on  MMLU~\cite{hendrycks2020measuring} and WinoGrande~\cite{sakaguchi2020winogrande}, 10-shot on HellaSwag~\cite{zellers2019hellaswag}, and 25-shot on ARC Challenge~\cite{clark2018think}.

\subsection{\method{} Proxy Validation: Performance Correlation and Stability}
\label{subsec:proxy_stability}
To evaluate \method{} as a proxy for recovery potential and verify its consistency across hyperparameters, we analyze 70 candidate models sampled from the AmoebaLLM~\cite{fu2024amoeballm} search space. This evaluates the proxy in an environment unbiased by our specific evolutionary search or large-scale CPT. We compute Spearman Rho (SP) and Kendall Tau (KT) correlations against Wikitext-2 perplexity (PPL), MMLU accuracy and average downstream accuracy after Once-for-All (OFA)~\cite{cai2019once,sukthanker2024large} finetuning on Alpaca~\cite{taori2023stanford}.

\begin{table}[ht]
\centering
\caption{Correlation of zero-shot proxies with model performance calculated over 70 pruned models and averaged over 3 seeds. We report Spearman Rho (SP) and Kendall Tau (KT) for perplexity (PPL), MMLU and average commonsense reasoning accuracies. Best correlation values per column are \textbf{bolded} and \underline{underlined} values denote second best correlation.}
\label{tab:proxy_correlation}
\footnotesize
\renewcommand{\arraystretch}{0.9} 
\setlength{\tabcolsep}{3pt} 
\begin{tabular}{l cc cc cc}
\toprule
\multirow{2}{*}{Method} & \multicolumn{2}{c}{PPL} & \multicolumn{2}{c}{MMLU} & \multicolumn{2}{c}{Avg. Acc.} \\
\cmidrule(lr){2-3} \cmidrule(lr){4-5} \cmidrule(lr){6-7}
& SP & KT & SP & KT & SP & KT \\
\midrule
\#Params & 0.87 & 0.72 & 0.11 & 0.07 & 0.66 & 0.50 \\
NASWOT & 0.69 & 0.50 & 0.35 & 0.26 & 0.39 & 0.29 \\
ZiCo & 0.87 & 0.72 & 0.11 & 0.07 & 0.66 & 0.50 \\
GradNorm & \textbf{0.93} & \textbf{0.79} & 0.45 & 0.32 & \underline{0.93} & \underline{0.78} \\
Synaptic Saliency & \textbf{0.93} & 0.78 & 0.37 & 0.25 & 0.89 & 0.73 \\
MeCo & 0.92 & \textbf{0.79} & 0.03 & 0.02 & 0.75 & 0.58 \\
PrunerZero & 0.44 & 0.31 & \textbf{0.88} & \textbf{0.69} & 0.68 & 0.50 \\
\midrule
\method{} -- Dot & 0.57 & 0.46 & 0.35 & 0.25 & 0.58 & 0.47 \\
\method{} -- Cosine & 0.91 & 0.77 & 0.48 & 0.35 & 0.91 & 0.79 \\
\method{} -- Unweighted & 0.85 & 0.71 & 0.23 & 0.16 & 0.77 & 0.62 \\
\method{} (Ours) & \textbf{0.93} & \textbf{0.79} & \underline{0.54} & \underline{0.39} & \textbf{0.94} & \textbf{0.82} \\
\bottomrule
\end{tabular}
\end{table}
\paragraph{Correlation with Downstream Performance} \method{} achieves superior ranking correlation with downstream performance, as shown in Table \ref{tab:proxy_correlation}. Unlike trainability proxies like NASWOT~\cite{mellor2021neural} and ZiCo~\cite{li2023zico}, which rank models based on their expressivity and convergence capability from random initializations, our proxy explicitly accounts for pretrained knowledge inheritance. By measuring the alignment between the pruned and base model gradient traces, \method{} captures how structural changes impact the functional sensitivity of the model. This allows the proxy to account for complex reasoning dependencies that feature-based or loss-landscape smoothness-based metrics fail to capture. 

 While GradNorm~\cite{abdelfattah2021zero} and Synaptic Saliency~\cite{tanaka2020pruning} effectively rank PPL, they prioritize the health of the gradient signal and whether pruning disproportionately impacts individual layers. This provides an incomplete picture of the candidate's representational integrity. Similarly, while MeCo~\cite{jiang2023meco} performs well on PPL, it fails to translate to task-specific accuracy. PrunerZero~\cite{dong2024pruner} achieves the highest correlation on MMLU by incorporating gradient information, but this success does not generalize to PPL or average downstream accuracy. Additionally, \#Params has high ranking potential for both PPL and average accuracy, however this does not translate to a good measure of a model's knowledge retention capability, which is needed for MMLU. This tells us that while increasing model size is a good indicator of generalizability potential, \#Params cannot capture the nuances required to analyze a pruned model's factual knowledge and reasoning capabilities. 

 In contrast, \method{} maintains superior ranking stability across all three metrics by leveraging Sparsity-Weighted Pearson Correlation $(\Phi)$. $\Phi$ is specifically designed to isolate functional signals from directional noise through gradient centering. By using layer-wise sparsity as a weighting factor, $\Phi$ anchors the global fitness score in high-capacity regions. This prevents the ranking from being skewed by the high variance noise typical of aggressively pruned sub-blocks. We further validate the choice of our proxy by comparing against alternative alignment metrics, including dot product, cosine similarity and unweighted aggregation of Pearson Correlation Coefficients. A detailed analysis of the correlation results is presented in Table~\ref{tab:proxy_correlation} is provided in Appendix~\ref{appendix:downstream_correlation_analysis}.

\begin{figure}[t]
    \centering
    \includegraphics[width=0.85\columnwidth]{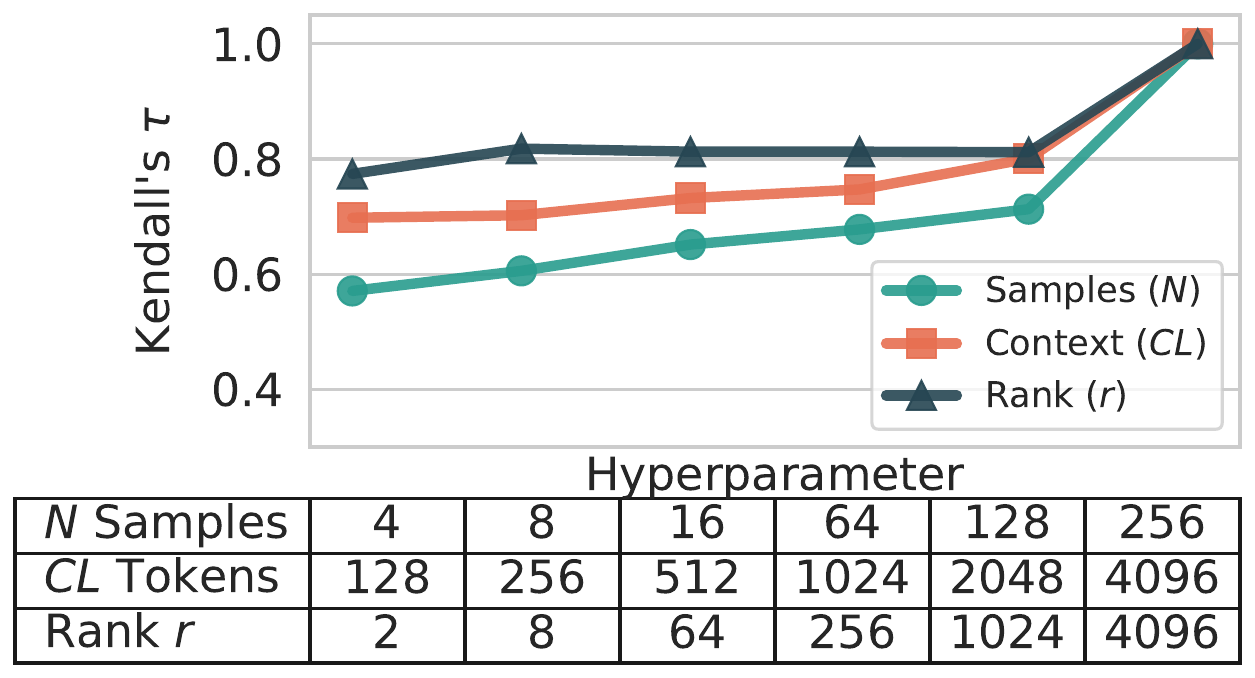}
    \caption{\textbf{\method{} proxy stability analysis.} We report Kendall $\tau$ correlation between ranking scores across search hyperparameters (a) number of samples (N), (b) context length (CL), and (c) LoRA rank ($r$) which define the x-axis. High $\tau$ values demonstrate that \method{} consistently ranks models across search settings.}
    \label{fig:stability_triple}
    \vspace{-3ex}
\end{figure}
\paragraph{Hyperparameter Stability Analysis} We evaluate the robustness of \method{} by computing the Kendall $\tau$ correlation between various search hyperparameters and the highest setting in each category ($N=256$, $CL=4096$, $r=4096$). As shown in Fig. \ref{fig:stability_triple}, the proxy maintains high ranking consistency across all variables, confirming that gradient trace alignment effectively captures model inheritance even under constrained search settings. Our analysis of these internal ranking correlations follows:
\begin{itemize}[itemsep=1pt, parsep=0pt, topsep=1pt] 
    \item \textbf{Number of Calibration Samples ($N$):} We observe that correlation monotonically align with the $N=256$ configuration as sample density increases. While ranking sensitivity is present at $N=4$ with $\tau \approx 0.57$, stability improves significantly to $\tau \approx 0.71$ at $N=128$. This trend suggests that representative gradient alignment is captured with relatively few samples, justifying our sample-efficient search settings.
    \item \textbf{Context Length ($CL$):} We observe that correlation with $CL=4096$ remains high across context lengths, rising from $\tau \approx 0.70$ at shorter contexts to $\tau \approx 0.80$ at $CL=2048$. The high agreement for all $CL \geq 1024$ rankings suggests that the proxy's saliency is largely preserved once the model captures sufficient long-range dependencies, allowing for significantly reduced compute during pruned model discovery.
    \item \textbf{Low-rank Gradient Subspace ($r$):} We observe that proxy stability is most significant across varying gradient subspace dimensions. Correlation with the full-rank $(r=4096)$ configuration remains consistently high, exceeding $\tau \approx 0.77$ even at $r=2$, and peaking at $\tau \approx 0.82$ for $r=8$ before stabilizing. This validates our use of $r=64$ during the search, confirming that low-rank gradient traces are sufficient to capture functional nuances while minimizing memory overhead during the search.
\end{itemize}
Detailed inter-hyperparameter correlations, comparing each configuration directly to downstream task performance, are provided in Appendix~\ref{appendix:corr_amoaballm_stability}.

\begin{table*}[t]
\centering
\begin{threeparttable}
\caption{Pruning results for LLaMA-2-7B. Averages are calculated across eight reasoning benchmarks. \method{} achieves highest average accuracy while requiring significantly fewer search tokens.}
\label{tab:llama2_results}
\vspace{-1ex}
\footnotesize
\renewcommand{\arraystretch}{0.9} 
\setlength{\tabcolsep}{5pt} 
\begin{tabular}{l c cc | ccccc ccc c}
\toprule
Method & \#Params & Srch & Recv & PIQA & WG & ArcE & ArcC & HS & SciQ & LQA & BQ & Avg. \\
\midrule
Llama-2-7B & 6.7B & -- & -- & 77.69 & 74.11 & 76.34 & 52.81 & 78.96 & 93.80 & 29.80 & 77.73 & 70.16 \\
\midrule
LoRAP & 2.7B & -- & -- & 57.20 & 47.90 & 31.30 & 26.30 & 30.00 & 51.20 & 27.50 & 61.90 & 41.66 \\
Flextron & 3.4B & 1K & 90B & 74.10 & 62.00 & 66.50 & -- & 68.50 & -- & -- & -- & -- \\ 
$\text{E}^3$ & 2.7B & 0.5B & -- & 71.20 & 57.30 & 65.40 & 38.10 & 54.60 & 88.40 & 28.10 & 63.60 & 58.34 \\
ShearedLLaMA & 2.7B & 0.4B & 50B & \textbf{75.80} & \textbf{64.20} & 67.00 & 41.20 & \textbf{70.80} & 90.80 & 28.20 & 63.00 & 62.63 \\
DarwinLM & 2.7B & 1.0B & 10B & 73.12 & 63.77 & 69.94 & \textbf{43.94} & 66.95 & 90.20 & 28.57 & \textbf{64.10} & 62.57 \\
\midrule
Uniform & 3.4B & 98M & 20B & 65.10 & 58.30 & 69.50 & 38.20 & 54.30 & 87.10 & 26.03 & 59.90 & 57.30 \\
ShearedLLaMA & 2.7B & 0.4B & 20B\tnote{$\dagger$} & 72.92 & 60.80 & 67.46 & 38.06 & 64.73 & 90.30 & 28.72 & 63.91 & 60.86 \\
DarwinLM & 2.7B & 1.0B & 20B$^\dagger$ & 73.17 & 63.59 & 69.90 & 42.91 & 66.57 & 89.00 & 29.18 & 63.63 & 62.24 \\
\method{} (Ours) & 2.7B & 98M & 20B & 73.40 & 63.97 & \textbf{70.20} & 43.80 & 68.54 & \textbf{91.10} & \textbf{29.71} & 61.76 & \textbf{62.81} \\
\bottomrule
\end{tabular}
\end{threeparttable}
\end{table*}
\begin{table*}[t]
\centering
\caption{Pruning results for Llama-3.1-8B and Qwen-2.5-14B-Instruct. \method{} has the best average accuracy and surpasses baselines on several complex reasoning benchmarks. Averages are calculated across the eight reasoning benchmarks, not including MMLU. }
\label{tab:combined_results}
\vspace{-1ex}
\footnotesize
\renewcommand{\arraystretch}{0.9} 
\setlength{\tabcolsep}{5pt} 
\begin{tabular}{l l cc | c ccccc ccc c}
\toprule
Method & \#Params & Srch & Recv & MMLU  & PIQA & WG & ArcE & ArcC & HS & SciQ & LQA & BQ & Avg. \\
\midrule
Llama-3.1-8B & 8B & -- & -- & 64.78 & 81.22 & 77.42 & 81.64 & 57.59 & 81.81 & 96.20 & 31.02 & 82.14 & 73.63 \\
\midrule
Minitron & 4.4B & -- & 94B & 60.50 & 76.82 & 73.50 & 76.89 & 56.14 & 76.03 & 96.20 & 31.02 & 83.14 & 71.22 \\
DarwinLM & 4.6B & 1B & 10B & 43.21 & \textbf{74.59} & 65.03 & 73.27 & 51.27 & 71.13 & 93.40 & 30.72 & 71.00 & 66.30 \\
Uniform (Ours) & 4.6B & 98M & 20B & 28.76 & 68.32 & 58.54 & 68.23 & 47.89 & 66.42 & 88.48 & 27.54 & 66.59 & 61.50 \\
DarwinLM & 4.6B & 1B & 20B$^\dagger$ & \textbf{43.32} & 74.53 & 65.26 & 73.13 & 51.62 & \textbf{71.17} & 93.24 & 30.76 & 70.08 & 66.22 \\
\method{} (Ours) & 4.6B & 98M & 20B & 30.12 & 74.23 & \textbf{66.08} & \textbf{73.50} & \textbf{52.26} & 69.01 & \textbf{93.90} & \textbf{30.81} & \textbf{73.00} & \textbf{66.60} \\
\midrule
\midrule
Qwen-2.5 14B & 14B & -- & -- & 82.53 & 82.10 & 79.71 & 85.81 & 72.44 & 85.10 & 96.70 & 41.16 & 87.95 & 78.87 \\
\midrule
Minitron & 8.4B & -- & 94B & 36.10 & 80.41 & 80.03 & 83.38 & 64.24 & 83.10 & 97.10 & 33.48 & 84.43 & 75.77 \\
DarwinLM & 8.4B & 1B & 10B & 55.97 & 78.12 & 70.63 & 79.41 & \textbf{57.42} & 74.93 & 89.30 & 33.10 & 73.57 & 69.56 \\
$\text{E}^3$ & 8.2B & 0.5B & -- & 36.50 & 76.90 & 63.00 & 76.00 & 47.90 & 67.00 & 93.70 & 30.00 & 66.50 & 65.13 \\
Uniform (Ours) & 8.4B & 98M & 20B & 51.23 & 71.32 & 62.64 & 77.12 & 50.02 & 69.89 & 88.91 & 28.53 & 69.45 & 64.73 \\
DarwinLM & 8.4B & 1B & 20B$^\dagger$ & 56.97 & \textbf{78.17} & 72.09 & 79.86 & 57.03 & \textbf{74.96} & 90.32 & \textbf{32.80} & 73.17 & 69.80 \\
\method{} (Ours) & 8.4B & 98M & 20B & \textbf{58.79} & 76.63 & \textbf{73.24} & \textbf{81.07} & 56.05 & 73.77 & \textbf{95.70} & 32.56 & \textbf{73.70} & \textbf{70.34} \\
\bottomrule
\end{tabular} 
\vspace{-3ex}
\end{table*}
\subsection{Main Results}
We evaluate the capability of \method{} to discover high performing pruned models by evaluating on Llama-2-7B, Llama-3.1-8B, and Qwen-2.5-14B Instruct to 2.7B, 4.6B, and 8.4B parameters, respectively. As shown in Tables~\ref{tab:llama2_results} and~\ref{tab:combined_results}, \method{} identifies sub-networks with high sample efficiency, requiring only 98M total search tokens, a $10\times$ and $4\times$ reduction compared to DarwinLM and ShearedLLaMA, while achieving superior accuracy. This efficiency validates the high ranking capability and stability of our proxy observed in Sec.~\ref{subsec:proxy_stability}.

\paragraph{Performance on Llama-2-7B} \method{} achieves the highest average accuracy (62.81\%) on Llama-2 7B across eight benchmarks, outperforming DarwinLM (62.57\%) and ShearedLLaMA (62.63\%). To ensure a fair comparison, we retrained DarwinLM and ShearedLLaMA on our 20B token distribution $(\dagger)$. Notably, ShearedLLaMA's performance drops to 60.86\% when restricted to our 20B token budget, highlighting its reliance on high-token-count recovery to reach competitive accuracy. While DarwinLM remains competitive, its performance slightly dips when using our unfiltered FineWeb-Edu samples rather than the highly curated top 0.1 percentile used in its original report.

Furthermore, \method{} significantly outperforms LoRAP~\cite{li2024lorap}, E$^3$-Pruner~\cite{yuan20253}, and our uniform baseline. Our 2.7B model has comparable the performance with Flextron~\cite{cai2024flextron} despite the latter having a larger parameter footprint and requiring $4.5\times$ more CPT tokens. The substantial performance gap between \method{} and the uniform baseline (5.51\% average) underscores the necessity of non-uniform architecture search for maintaining representational integrity. We provide performance scalability results of our Llama-2-2.7B model by evaluating recovery performance at 10B and 50B tokens in Appendix~\ref{appendix:token_scales}.

{
  \renewcommand{\thefootnote}{\fnsymbol{footnote}}
  \footnotetext[2]{Models trained on our CPT data setup and the original codebase provided in the respective papers.}
}
\paragraph{Generalization to GQA Architectures} Evaluation on Llama-3.1 and Qwen-2.5 demonstrates the generalizability of \method{} to GQA architectures. For Llama-3.1-8B, \method{} significantly improves upon the uniform baseline (66.60\% vs 61.50\%) and outperforms DarwinLM across critical reasoning tasks such as ARC-C (52.26\%) and BoolQ (73.00\%). On Qwen-2.5-14B, \method{} maintains its lead with a 70.34\% average, surpassing both DarwinLM (69.80\%) and E$^3$-Pruner (65.13\%).

While Minitron~\cite{sreenivas2024llm} achieves higher accuracies across reasoning baseline and \method{} models, it requires an intensive 94B tokens for recovery, approximately $4.7\times$ the computational cost of \method{}. Across all architectures, \method{} produces sub-networks whose gradient traces are highly aligned with the pretrained gradient trace, allowing for high-fidelity recovery and discovered at a fraction of the computational search cost required by existing training-aware methods. Detailed perplexity evaluations across different sparsity levels and a speedup analysis against all publicly available baselines are provided in Appendix~\ref{appendix:ppl_analysis} and Appendix~\ref{appendix:speedup_analysis}, respectively.
\section{Conclusion}

In this work, we introduce \method, a zero-shot NAS framework for non-uniform structured pruning of LLMs. By leveraging a novel scale-invariant gradient proxy, \method{} identifies sub-networks that maintain high gradient trace alignment with the pretrained base model. This alignment ensures the preservation of functional inheritance, allowing for efficient recovery during continued pretraining. Our framework demonstrates consistent performance gains across Llama-2, Llama-3.1, and Qwen-2.5 architectures, achieving an order-of-magnitude reduction in search overhead compared to training-aware baselines. By eliminating the prohibitive cost of search-time training, \method{} provides a scalable and high-fidelity foundation for model compression that effectively retains the complex reasoning capabilities of dense pretrained~models.
\newpage
\newpage
\section*{Acknowledgments}
This work was supported in part by the Center for the Co-Design of Cognitive Systems (COCOSYS), a DARPA-sponsored JUMP center, the Semiconductor Research Corporation (SRC), the National Science Foundation (NSF) and the Department of Energy (DOE). 

\section*{Impact Statement}

The research presented in this paper advances the accessibility of high-performing LLMs by significantly lowering the computational barriers to large-scale model compression. By providing a training-free framework for architectural discovery, we enable the development of efficient models even under resource constraints. Furthermore, the substantial reduction in GPU-hours required for model search directly mitigates the carbon footprint and energy consumption associated with large-scale NAS. These advancements enable the deployment of LLMs in edge environments, supporting sustainable AI development. To promote reproducibility, an anonymous GitHub repository will be made available to reviewers and area chairs during the discussion period, as per the ICML Author Guidelines.
\bibliography{main}

@article{openai2023gpt,
  title={Gpt-4 technical report. arxiv 2303.08774},
  author={OpenAI, R},
  journal={View in Article},
  volume={2},
  number={5},
  pages={1},
  year={2023}
}

@article{anthropic2023introducing,
  title={Introducing claude},
  author={Anthropic, PBC},
  journal={March},
  volume={14},
  pages={2023},
  year={2023}
}

@article{touvron2023llama,
  title={Llama: Open and efficient foundation language models},
  author={Touvron, Hugo and Lavril, Thibaut and Izacard, Gautier and Martinet, Xavier and Lachaux, Marie-Anne and Lacroix, Timoth{\'e}e and Rozi{\`e}re, Baptiste and Goyal, Naman and Hambro, Eric and Azhar, Faisal and others},
  journal={arXiv preprint arXiv:2302.13971},
  year={2023}
}

@article{dubey2024llama,
  title={The llama 3 herd of models},
  author={Dubey, Abhimanyu and Jauhri, Abhinav and Pandey, Abhinav and Kadian, Abhishek and Al-Dahle, Ahmad and Letman, Aiesha and Mathur, Akhil and Yang, Amy and Fan, Angela and others},
  journal={arXiv preprint arXiv:2407.21783},
  year={2024}
}

@article{kaplan2020scaling,
  title={Scaling laws for neural language models},
  author={Kaplan, Jared and McCandlish, Sam and Henighan, Tom and Brown, Tom B and Chess, Benjamin and Child, Rewon and Gray, Scott and Radford, Alec and Wu, Jeffrey and Amodei, Dario},
  journal={arXiv preprint arXiv:2001.08361},
  year={2020}
}

@article{frantar2022gptq,
  title={Gptq: Accurate post-training quantization for generative pre-trained transformers},
  author={Frantar, Elias and Ashkboos, Saleh and Hoefler, Torsten and Alistarh, Dan},
  journal={arXiv preprint arXiv:2210.17323},
  year={2022}
}

@inproceedings{liu2024llm,
  title={Llm-qat: Data-free quantization aware training for large language models},
  author={Liu, Zechun and Oguz, Barlas and Zhao, Changsheng and Chang, Ernie and Stock, Pierre and Mehdad, Yashar and Shi, Yangyang and Krishnamoorthi, Raghuraman and Chandra, Vikas},
  booktitle={Findings of the Association for Computational Linguistics: ACL 2024},
  pages={467--484},
  year={2024}
}

@InProceedings{pmlr-v267-bercovich25a,
  title = 	 {Puzzle: Distillation-Based {NAS} for Inference-Optimized {LLM}s},
  author =       {Bercovich, Akhiad and Ronen, Tomer and Abramovich, Talor and Ailon, Nir and Assaf, Nave and Dabbah, Mohammed and Galil, Ido and Geifman, Amnon and Geifman, Yonatan and Golan, Izhak and Haber, Netanel and Karpas, Ehud Dov and Koren, Roi and Levy, Itay and Molchanov, Pavlo and Mor, Shahar and Moshe, Zach and Nabwani, Najeeb and Puny, Omri and Rubin, Ran and Schen, Itamar and Shahaf, Ido and Tropp, Oren and Argov, Omer Ullman and Zilberstein, Ran and El-Yaniv, Ran},
  booktitle = 	 {Proceedings of the 42nd International Conference on Machine Learning},
  pages = 	 {3806--3830},
  year = 	 {2025},
  editor = 	 {Singh, Aarti and Fazel, Maryam and Hsu, Daniel and Lacoste-Julien, Simon and Berkenkamp, Felix and Maharaj, Tegan and Wagstaff, Kiri and Zhu, Jerry},
  volume = 	 {267},
  series = 	 {Proceedings of Machine Learning Research},
  month = 	 {13--19 Jul},
  publisher =    {PMLR},
  url = 	 {https://proceedings.mlr.press/v267/bercovich25a.html}
}

@inproceedings{sarah2024llama,
  title={Llama-nas: Efficient neural architecture search for large language models},
  author={Sarah, Anthony and Nittur Sridhar, Sharath and Szankin, Maciej and Sundaresan, Sairam},
  booktitle={European Conference on Computer Vision},
  pages={67--74},
  year={2024},
  organization={Springer}
}

@article{li2022shadow,
  title={Shadow knowledge distillation: Bridging offline and online knowledge transfer},
  author={Li, Lujun and Jin, Zhe},
  journal={Advances in Neural Information Processing Systems},
  volume={35},
  pages={635--649},
  year={2022}
}

@article{han2015deep,
  title={Deep compression: Compressing deep neural networks with pruning, trained quantization and huffman coding},
  author={Han, Song and Mao, Huizi and Dally, William J},
  journal={arXiv preprint arXiv:1510.00149},
  year={2015}
}

@article{xia2023sheared,
  title={Sheared llama: Accelerating language model pre-training via structured pruning},
  author={Xia, Mengzhou and Gao, Tianyu and Zeng, Zhiyuan and Chen, Danqi},
  journal={arXiv preprint arXiv:2310.06694},
  year={2023}
}

@article{yuan20253,
  title={E $^3$-Pruner: Towards Efficient, Economical, and Effective Layer Pruning for Large Language Models},
  author={Yuan, Tao and Bai, Haoli and Pan, Yinfei and Cao, Xuyang and Zhang, Tianyu and Hou, Lu and Hu, Ting and Yu, Xianzhi},
  journal={arXiv preprint arXiv:2511.17205},
  year={2025}
}

@inproceedings{frantar2023sparsegpt,
  title={Sparsegpt: Massive language models can be accurately pruned in one-shot},
  author={Frantar, Elias and Alistarh, Dan},
  booktitle={International conference on machine learning},
  pages={10323--10337},
  year={2023},
  organization={PMLR}
}

@article{ma2023llm,
  title={Llm-pruner: On the structural pruning of large language models},
  author={Ma, Xinyin and Fang, Gongfan and Wang, Xinchao},
  journal={Advances in neural information processing systems},
  volume={36},
  pages={21702--21720},
  year={2023}
}

@article{sun2023simple,
  title={A simple and effective pruning approach for large language models},
  author={Sun, Mingjie and Liu, Zhuang and Bair, Anna and Kolter, J Zico},
  journal={arXiv preprint arXiv:2306.11695},
  year={2023}
}

@inproceedings{an2024fluctuation,
  title={Fluctuation-based adaptive structured pruning for large language models},
  author={An, Yongqi and Zhao, Xu and Yu, Tao and Tang, Ming and Wang, Jinqiao},
  booktitle={Proceedings of the AAAI Conference on Artificial Intelligence},
  volume={38},
  number={10},
  pages={10865--10873},
  year={2024}
}

@inproceedings{hassibi1993optimal,
  title={Optimal brain surgeon and general network pruning},
  author={Hassibi, Babak and Stork, David G and Wolff, Gregory J},
  booktitle={IEEE international conference on neural networks},
  pages={293--299},
  year={1993},
  organization={IEEE}
}

@article{li2024lorap,
  title={Lorap: Transformer sub-layers deserve differentiated structured compression for large language models},
  author={Li, Guangyan and Tang, Yongqiang and Zhang, Wensheng},
  journal={arXiv preprint arXiv:2404.09695},
  year={2024}
}

@article{sreenivas2024llm,
  title={Llm pruning and distillation in practice: The minitron approach},
  author={Sreenivas, Sharath Turuvekere and Muralidharan, Saurav and Joshi, Raviraj and Chochowski, Marcin and Mahabaleshwarkar, Ameya Sunil and Shen, Gerald and Zeng, Jiaqi and Chen, Zijia and Suhara, Yoshi and Diao, Shizhe and others},
  journal={arXiv preprint arXiv:2408.11796},
  year={2024}
}

@article{tang2025darwinlm,
  title={Darwinlm: Evolutionary structured pruning of large language models},
  author={Tang, Shengkun and Sieberling, Oliver and Kurtic, Eldar and Shen, Zhiqiang and Alistarh, Dan},
  journal={arXiv preprint arXiv:2502.07780},
  year={2025}
}

@article{dong2024pruner,
  title={Pruner-zero: Evolving symbolic pruning metric from scratch for large language models},
  author={Dong, Peijie and Li, Lujun and Tang, Zhenheng and Liu, Xiang and Pan, Xinglin and Wang, Qiang and Chu, Xiaowen},
  journal={arXiv preprint arXiv:2406.02924},
  year={2024}
}

@article{dong2024lpzero,
  title={Lpzero: Language model zero-cost proxy search from zero},
  author={Dong, Peijie and Li, Lujun and Liu, Xiang and Tang, Zhenheng and Liu, Xuebo and Wang, Qiang and Chu, Xiaowen},
  journal={arXiv preprint arXiv:2410.04808},
  year={2024}
}

@inproceedings{wang2020structured,
  title={Structured pruning of large language models},
  author={Wang, Ziheng and Wohlwend, Jeremy and Lei, Tao},
  booktitle={Proceedings of the 2020 conference on empirical methods in natural language processing (emnlp)},
  pages={6151--6162},
  year={2020}
}

@article{frankle2018lottery,
  title={The lottery ticket hypothesis: Finding sparse, trainable neural networks},
  author={Frankle, Jonathan and Carbin, Michael},
  journal={arXiv preprint arXiv:1803.03635},
  year={2018}
}

@article{chen2020lottery,
  title={The lottery ticket hypothesis for pre-trained bert networks},
  author={Chen, Tianlong and Frankle, Jonathan and Chang, Shiyu and Liu, Sijia and Zhang, Yang and Wang, Zhangyang and Carbin, Michael},
  journal={Advances in neural information processing systems},
  volume={33},
  pages={15834--15846},
  year={2020}
}

@inproceedings{frankle2020linear,
  title={Linear mode connectivity and the lottery ticket hypothesis},
  author={Frankle, Jonathan and Dziugaite, Gintare Karolina and Roy, Daniel and Carbin, Michael},
  booktitle={International Conference on Machine Learning},
  pages={3259--3269},
  year={2020},
  organization={PMLR}
}

@inproceedings{yu2020bignas,
  title={Bignas: Scaling up neural architecture search with big single-stage models},
  author={Yu, Jiahui and Jin, Pengchong and Liu, Hanxiao and Bender, Gabriel and Kindermans, Pieter-Jan and Tan, Mingxing and Huang, Thomas and Song, Xiaodan and Pang, Ruoming and Le, Quoc},
  booktitle={European Conference on Computer Vision},
  pages={702--717},
  year={2020},
  organization={Springer}
}

@article{li2018visualizing,
  title={Visualizing the loss landscape of neural nets},
  author={Li, Hao and Xu, Zheng and Taylor, Gavin and Studer, Christoph and Goldstein, Tom},
  journal={Advances in neural information processing systems},
  volume={31},
  year={2018}
}

@article{hu2022lora,
  title={Lora: Low-rank adaptation of large language models.},
  author={Hu, Edward J and Shen, Yelong and Wallis, Phillip and Allen-Zhu, Zeyuan and Li, Yuanzhi and Wang, Shean and Wang, Lu and Chen, Weizhu and others},
  journal={ICLR},
  volume={1},
  number={2},
  pages={3},
  year={2022}
}

@article{askari2025layerif,
  title={LayerIF: Estimating Layer Quality for Large Language Models using Influence Functions},
  author={Askari, Hadi and Gupta, Shivanshu and Wang, Fei and Chhabra, Anshuman and Chen, Muhao},
  journal={arXiv preprint arXiv:2505.23811},
  year={2025}
}

@article{kwon2023datainf,
  title={Datainf: Efficiently estimating data influence in lora-tuned llms and diffusion models},
  author={Kwon, Yongchan and Wu, Eric and Wu, Kevin and Zou, James},
  journal={arXiv preprint arXiv:2310.00902},
  year={2023}
}

@article{fu2024amoeballm,
  title={Amoeballm: Constructing any-shape large language models for efficient and instant deployment},
  author={Fu, Yonggan and Yu, Zhongzhi and Li, Junwei and Qian, Jiayi and Zhang, Yongan and Yuan, Xiangchi and Shi, Dachuan and Yakunin, Roman and Lin, Yingyan Celine},
  journal={Advances in Neural Information Processing Systems},
  volume={37},
  pages={78299--78319},
  year={2024}
}

@article{kim2024shortened,
  title={Shortened llama: A simple depth pruning for large language models},
  author={Kim, Bo-Kyeong and Kim, Geonmin and Kim, Tae-Ho and Castells, Thibault and Choi, Shinkook and Shin, Junho and Song, Hyoung-Kyu},
  journal={arXiv preprint arXiv:2402.02834},
  volume={11},
  pages={1},
  year={2024}
}

@article{cai2019once,
  title={Once-for-all: Train one network and specialize it for efficient deployment},
  author={Cai, Han and Gan, Chuang and Wang, Tianzhe and Zhang, Zhekai and Han, Song},
  journal={arXiv preprint arXiv:1908.09791},
  year={2019}
}

@inproceedings{sukthanker2024large,
  title={Large language model compression with neural architecture search},
  author={Sukthanker, Rhea Sanjay and Staffler, Benedikt and Hutter, Frank and Klein, Aaron},
  booktitle={Workshop on Machine Learning and Compression, NeurIPS 2024}
}

@article{cai2024flextron,
  title={Flextron: Many-in-one flexible large language model},
  author={Cai, Ruisi and Muralidharan, Saurav and Heinrich, Greg and Yin, Hongxu and Wang, Zhangyang and Kautz, Jan and Molchanov, Pavlo},
  journal={arXiv preprint arXiv:2406.10260},
  year={2024}
}

@article{das2023beyond,
  title={Beyond size: How gradients shape pruning decisions in large language models},
  author={Das, Rocktim Jyoti and Sun, Mingjie and Ma, Liqun and Shen, Zhiqiang},
  journal={arXiv preprint arXiv:2311.04902},
  year={2023}
}

@article{yang2025wanda++,
  title={Wanda++: Pruning large language models via regional gradients},
  author={Yang, Yifan and Zhen, Kai and Ganesh, Bhavana and Galstyan, Aram and Huybrechts, Goeric and M{\"u}ller, Markus and K{\"u}bler, Jonas M and Swaminathan, Rupak Vignesh and Mouchtaris, Athanasios and Bodapati, Sravan Babu and others},
  journal={arXiv preprint arXiv:2503.04992},
  year={2025}
}

@inproceedings{wu2021training,
  title={A training-free genetic neural architecture search},
  author={Wu, Meng-Ting and Lin, Hung-I and Tsai, Chun-Wei},
  booktitle={Proceedings of the 2021 ACM International Conference on Intelligent Computing and Its Emerging Applications},
  pages={65--70},
  year={2021}
}

@inproceedings{ingolfsson2022reducing,
  title={Reducing neural architecture search spaces with training-free statistics and computational graph clustering},
  author={Ingolfsson, Thorir Mar and Vero, Mark and Wang, Xiaying and Lamberti, Lorenzo and Benini, Luca and Spallanzani, Matteo},
  booktitle={Proceedings of the 19th ACM International Conference on Computing Frontiers},
  pages={213--214},
  year={2022}
}

@article{tran2021feature,
  title={A feature fusion based indicator for training-free neural architecture search},
  author={Tran, Linh-Tam and Ali, Muhammad Salman and Bae, Sung-Ho},
  journal={IEEE Access},
  volume={9},
  pages={133914--133923},
  year={2021},
  publisher={IEEE}
}

@article{li2023zico,
  title={Zico: Zero-shot nas via inverse coefficient of variation on gradients},
  author={Li, Guihong and Yang, Yuedong and Bhardwaj, Kartikeya and Marculescu, Radu},
  journal={arXiv preprint arXiv:2301.11300},
  year={2023}
}

@inproceedings{mellor2021neural,
  title={Neural architecture search without training},
  author={Mellor, Joe and Turner, Jack and Storkey, Amos and Crowley, Elliot J},
  booktitle={International conference on machine learning},
  pages={7588--7598},
  year={2021},
  organization={PMLR}
}

@article{tanaka2020pruning,
  title={Pruning neural networks without any data by iteratively conserving synaptic flow},
  author={Tanaka, Hidenori and Kunin, Daniel and Yamins, Daniel L and Ganguli, Surya},
  journal={Advances in neural information processing systems},
  volume={33},
  pages={6377--6389},
  year={2020}
}

@article{jiang2023meco,
  title={Meco: zero-shot NAS with one data and single forward pass via minimum eigenvalue of correlation},
  author={Jiang, Tangyu and Wang, Haodi and Bie, Rongfang},
  journal={Advances in Neural Information Processing Systems},
  volume={36},
  pages={61020--61047},
  year={2023}
}

@article{abdelfattah2021zero,
  title={Zero-cost proxies for lightweight NAS},
  author={Abdelfattah, Mohamed S and Mehrotra, Abhinav and Dudziak, {\L}ukasz and Lane, Nicholas D},
  journal={arXiv preprint arXiv:2101.08134},
  year={2021}
}

@article{gromov2024unreasonable,
  title={The unreasonable ineffectiveness of the deeper layers, 2024},
  author={Gromov, Andrey and Tirumala, Kushal and Shapourian, Hassan and Glorioso, Paolo and Roberts, Daniel A},
  journal={URL https://arxiv. org/abs/2403.17887},
  year={2024}
}

@article{vaswani2017attention,
  title={Attention is all you need},
  author={Vaswani, Ashish and Shazeer, Noam and Parmar, Niki and Uszkoreit, Jakob and Jones, Llion and Gomez, Aidan N and Kaiser, {\L}ukasz and Polosukhin, Illia},
  journal={Advances in neural information processing systems},
  volume={30},
  year={2017}
}

@article{ainslie2023gqa,
  title={Gqa: Training generalized multi-query transformer models from multi-head checkpoints},
  author={Ainslie, Joshua and Lee-Thorp, James and De Jong, Michiel and Zemlyanskiy, Yury and Lebr{\'o}n, Federico and Sanghai, Sumit},
  journal={arXiv preprint arXiv:2305.13245},
  year={2023}
}

@inproceedings{aghajanyan-etal-2021-intrinsic,
    title = "Intrinsic Dimensionality Explains the Effectiveness of Language Model Fine-Tuning",
    author = "Aghajanyan, Armen  and
      Gupta, Sonal  and
      Zettlemoyer, Luke",
    editor = "Zong, Chengqing  and
      Xia, Fei  and
      Li, Wenjie  and
      Navigli, Roberto",
    booktitle = "Proceedings of the 59th Annual Meeting of the Association for Computational Linguistics and the 11th International Joint Conference on Natural Language Processing (Volume 1: Long Papers)",
    month = aug,
    year = "2021",
    address = "Online",
    publisher = "Association for Computational Linguistics",
    url = "https://aclanthology.org/2021.acl-long.568/",
    doi = "10.18653/v1/2021.acl-long.568",
    pages = "7319--7328"
}

@article{zhao2024galore,
  title={Galore: Memory-efficient llm training by gradient low-rank projection},
  author={Zhao, Jiawei and Zhang, Zhenyu and Chen, Beidi and Wang, Zhangyang and Anandkumar, Anima and Tian, Yuandong},
  journal={arXiv preprint arXiv:2403.03507},
  year={2024}
}

@article{penedo2024fineweb,
  title={The fineweb datasets: Decanting the web for the finest text data at scale},
  author={Penedo, Guilherme and Kydl{\'\i}{\v{c}}ek, Hynek and Lozhkov, Anton and Mitchell, Margaret and Raffel, Colin A and Von Werra, Leandro and Wolf, Thomas and others},
  journal={Advances in Neural Information Processing Systems},
  volume={37},
  pages={30811--30849},
  year={2024}
}

@article{mukherjee2023orca,
  title={Orca: Progressive learning from complex explanation traces of gpt-4},
  author={Mukherjee, Subhabrata and Mitra, Arindam and Jawahar, Ganesh and Agarwal, Sahaj and Palangi, Hamid and Awadallah, Ahmed},
  journal={arXiv preprint arXiv:2306.02707},
  year={2023}
}

@misc{taori2023stanford,
  title={Stanford alpaca: An instruction-following llama model},
  author={Taori, Rohan and Gulrajani, Ishaan and Zhang, Tianyi and Dubois, Yann and Li, Xuechen and Guestrin, Carlos and Liang, Percy and Hashimoto, Tatsunori B},
  year={2023},
  publisher={Stanford, CA, USA}
}

@article{clark2018think,
  title={Think you have solved question answering? try arc, the ai2 reasoning challenge},
  author={Clark, Peter and Cowhey, Isaac and Etzioni, Oren and Khot, Tushar and Sabharwal, Ashish and Schoenick, Carissa and Tafjord, Oyvind},
  journal={arXiv preprint arXiv:1803.05457},
  year={2018}
}

@article{liu2020logiqa,
  title={Logiqa: A challenge dataset for machine reading comprehension with logical reasoning},
  author={Liu, Jian and Cui, Leyang and Liu, Hanmeng and Huang, Dandan and Wang, Yile and Zhang, Yue},
  journal={arXiv preprint arXiv:2007.08124},
  year={2020}
}

@inproceedings{bisk2020piqa,
  title={Piqa: Reasoning about physical commonsense in natural language},
  author={Bisk, Yonatan and Zellers, Rowan and Gao, Jianfeng and Choi, Yejin and others},
  booktitle={Proceedings of the AAAI conference on artificial intelligence},
  volume={34},
  number={05},
  pages={7432--7439},
  year={2020}
}

@article{welbl2017crowdsourcing,
  title={Crowdsourcing multiple choice science questions},
  author={Welbl, Johannes and Liu, Nelson F and Gardner, Matt},
  journal={arXiv preprint arXiv:1707.06209},
  year={2017}
}

@article{clark2019boolq,
  title={Boolq: Exploring the surprising difficulty of natural yes/no questions},
  author={Clark, Christopher and Lee, Kenton and Chang, Ming-Wei and Kwiatkowski, Tom and Collins, Michael and Toutanova, Kristina},
  journal={arXiv preprint arXiv:1905.10044},
  year={2019}
}

@article{hendrycks2020measuring,
  title={Measuring massive multitask language understanding},
  author={Hendrycks, Dan and Burns, Collin and Basart, Steven and Zou, Andy and Mazeika, Mantas and Song, Dawn and Steinhardt, Jacob},
  journal={arXiv preprint arXiv:2009.03300},
  year={2020}
}

@inproceedings{sakaguchi2020winogrande,
  title={Winogrande: An adversarial winograd schema challenge at scale},
  author={Sakaguchi, Keisuke and Le Bras, Ronan and Bhagavatula, Chandra and Choi, Yejin},
  booktitle={Proceedings of the AAAI Conference on Artificial Intelligence},
  volume={34},
  number={05},
  pages={8732--8740},
  year={2020}
}

@article{zellers2019hellaswag,
  title={Hellaswag: Can a machine really finish your sentence?},
  author={Zellers, Rowan and Holtzman, Ari and Bisk, Yonatan and Farhadi, Ali and Choi, Yejin},
  journal={arXiv preprint arXiv:1905.07830},
  year={2019}
}

@article{kovaleva2021bert,
  title={BERT busters: Outlier dimensions that disrupt transformers},
  author={Kovaleva, Olga and Kulshreshtha, Saurabh and Rogers, Anna and Rumshisky, Anna},
  journal={arXiv preprint arXiv:2105.06990},
  year={2021}
}

@inproceedings{luo2021positional,
  title={Positional artefacts propagate through masked language model embeddings},
  author={Luo, Ziyang and Kulmizev, Artur and Mao, Xiaoxi},
  booktitle={Proceedings of the 59th annual meeting of the Association for Computational Linguistics and the 11th international joint conference on natural language processing (volume 1: long papers)},
  pages={5312--5327},
  year={2021}
}

@article{veit2016residual,
  title={Residual networks behave like ensembles of relatively shallow networks},
  author={Veit, Andreas and Wilber, Michael J and Belongie, Serge},
  journal={Advances in neural information processing systems},
  volume={29},
  year={2016}
}

@article{spearman1961proof,
  title={The proof and measurement of association between two things.},
  author={Spearman, Charles},
  year={1961},
  publisher={Appleton-Century-Crofts}
}

@article{kendall1938new,
  title={A new measure of rank correlation},
  author={Kendall, Maurice G},
  journal={Biometrika},
  volume={30},
  number={1-2},
  pages={81--93},
  year={1938},
  publisher={Oxford University Press}
}

@article{yin2023outlier,
  title={Outlier weighed layerwise sparsity (owl): A missing secret sauce for pruning llms to high sparsity},
  author={Yin, Lu and Wu, You and Zhang, Zhenyu and Hsieh, Cheng-Yu and Wang, Yaqing and Jia, Yiling and Li, Gen and Jaiswal, Ajay and Pechenizkiy, Mykola and Liang, Yi and others},
  journal={arXiv preprint arXiv:2310.05175},
  year={2023}
}

@inproceedings{tao-etal-2023-structured,
    title = "Structured Pruning for Efficient Generative Pre-trained Language Models",
    author = "Tao, Chaofan  and
      Hou, Lu  and
      Bai, Haoli  and
      Wei, Jiansheng  and
      Jiang, Xin  and
      Liu, Qun  and
      Luo, Ping  and
      Wong, Ngai",
    editor = "Rogers, Anna  and
      Boyd-Graber, Jordan  and
      Okazaki, Naoaki",
    booktitle = "Findings of the Association for Computational Linguistics: ACL 2023",
    month = jul,
    year = "2023",
    address = "Toronto, Canada",
    publisher = "Association for Computational Linguistics",
    url = "https://aclanthology.org/2023.findings-acl.692/",
    doi = "10.18653/v1/2023.findings-acl.692",
    pages = "10880--10895"
}

@inproceedings{sieberlingevopress,
  title={EvoPress: Accurate Dynamic Model Compression via Evolutionary Search},
  author={Sieberling, Oliver and Kuznedelev, Denis and Kurtic, Eldar and Alistarh, Dan},
  booktitle={Forty-second International Conference on Machine Learning}
}

@misc{yuan2025e3prunerefficienteconomicaleffective,
      title={E$^3$-Pruner: Towards Efficient, Economical, and Effective Layer Pruning for Large Language Models}, 
      author={Tao Yuan and Haoli Bai and Yinfei Pan and Xuyang Cao and Tianyu Zhang and Lu Hou and Ting Hu and Xianzhi Yu},
      year={2025},
      eprint={2511.17205},
      archivePrefix={arXiv},
      primaryClass={cs.CL},
      url={https://arxiv.org/abs/2511.17205}, 
}

@inproceedings{koh2017understanding,
  title={Understanding black-box predictions via influence functions},
  author={Koh, Pang Wei and Liang, Percy},
  booktitle={International conference on machine learning},
  pages={1885--1894},
  year={2017},
  organization={PMLR}
}

@article{pruthi2020estimating,
  title={Estimating training data influence by tracing gradient descent},
  author={Pruthi, Garima and Liu, Frederick and Kale, Satyen and Sundararajan, Mukund},
  journal={Advances in Neural Information Processing Systems},
  volume={33},
  pages={19920--19930},
  year={2020}
}

@article{hu2024minicpm,
  title={Minicpm: Unveiling the potential of small language models with scalable training strategies},
  author={Hu, Shengding and Tu, Yuge and Han, Xu and He, Chaoqun and Cui, Ganqu and Long, Xiang and Zheng, Zhi and Fang, Yewei and Huang, Yuxiang and Zhao, Weilin and others},
  journal={arXiv preprint arXiv:2404.06395},
  year={2024}
}

@misc{eval-harness,
  author       = {Gao, Leo and Tow, Jonathan and Abbasi, Baber and Biderman, Stella and Black, Sid and DiPofi, Anthony and Foster, Charles and Golding, Laurence and Hsu, Jeffrey and Le Noac'h, Alain and Li, Haonan and McDonell, Kyle and Muennighoff, Niklas and Ociepa, Chris and Phang, Jason and Reynolds, Laria and Schoelkopf, Hailey and Skowron, Aviya and Sutawika, Lintang and Tang, Eric and Thite, Anish and Wang, Ben and Wang, Kevin and Zou, Andy},
  title        = {The Language Model Evaluation Harness},
  month        = 07,
  year         = 2024,
  publisher    = {Zenodo},
  version      = {v0.4.3},
  doi          = {10.5281/zenodo.12608602},
  url          = {https://zenodo.org/records/12608602}
}

@article{tran2022pruning,
  title={Pruning has a disparate impact on model accuracy},
  author={Tran, Cuong and Fioretto, Ferdinando and Kim, Jung-Eun and Naidu, Rakshit},
  journal={Advances in neural information processing systems},
  volume={35},
  pages={17652--17664},
  year={2022}
}
\bibliographystyle{icml2026}
\newpage
\makeatletter
\def\addcontentsline#1#2#3{%
  \addtocontents{#1}{\protect\contentsline{#2}{#3}{\thepage}{\@currentHref}}%
}

\def\l@subsection#1#2{\@dottedtocline{2}{1.5em}{2.8em}{#1}{#2}}
\def\l@subsubsection#1#2{\@dottedtocline{3}{3.8em}{3.2em}{#1}{#2}}
\makeatother




\newpage
\appendix
\onecolumn
\raggedbottom 
\etocdepthtag.toc{atoc}
\section*{Appendix}

\newcommand{\appendixtoc}{%
    \begingroup
    \etocsettagdepth{mtoc}{none}
    \etocsettagdepth{atoc}{subsubsection}
    
    \parbox[b]{0.96\textwidth}{
        \etocsettocstyle{\subsubsection*{Table of Contents\\ \vspace{-0.75em}\rule{\textwidth}{0.4pt}}\vspace{-.75em}}{}%
        \etoctableofcontents 
        \vspace{-0.25em} 
        \rule{\textwidth}{0.4pt}
    }\\
    \endgroup
}
\appendixtoc

\section{Implementation \& Search Analysis}
\label{appendix:search_space_init}
\subsection{Evolutionary Search Initialization}
\begin{table*}[h]
\centering
\caption{Comparison of LLaMA-2-7B pruning using evolutionary search and importance based initialization, highlighted as \method{} (Ours).}
\label{tab:llama2_results_init}
\begin{footnotesize}
\renewcommand{\arraystretch}{0.7} 
\setlength{\tabcolsep}{5pt} 
\begin{tabular}{l c ccccc ccc c}
\toprule
Method & \#Params & PIQA & WG & ArcE & ArcC & HS & SciQ & LQA & BQ & Avg. \\
\midrule
Llama-2-7B & 6.7B & 77.69 & 74.11 & 76.34 & 52.81 & 78.96 & 93.80 & 29.80 & 77.73 & 70.16 \\
\midrule
\method{} - Random Search & 2.7B & 68.42 & 58.86 & 66.72 & 37.13 & 58.45 & 88.79 & 27.85 & 59.98 & 58.28 \\
\method{} - Uniform Init & 2.7B & 72.24 & 61.21 & 69.72 & 41.79 & 65.24 & 89.96 & 28.84 & 59.79 & 61.10 \\
\method{} - Random Init  & 2.7B & 71.96 & 60.87 & 68.98 & 40.75 & 63.11 & 90.75 & 28.76 & 59.85 & 60.63 \\
\method{} (Ours) & 2.7B & 73.40 & 63.97 & \textbf{70.20} & 43.80 & 68.54 & \textbf{91.10} & \textbf{29.71} & 61.76 & \textbf{62.81} \\
\bottomrule
\end{tabular}
\end{footnotesize}
\end{table*}
To improve search efficiency and avoid sub-optimal architectural configurations, we warm-start the evolutionary search using a layer importance global prior. Instead of initializing the population uniformly or randomly within the search space, we bias candidates toward structurally stable regions of the pretrained model. Specifically, we compute per-layer importance scores based on the expectation of the weight-activation product~\cite{sun2023simple}, defined as $I_l = \mathbb{E}\big[ |W_l| \odot \|X_l\|_2 \big]$, which characterizes the sensitivity of each layer.

These importance scores define a non-uniform sampling prior over the sub-module width sparsity configurations $\boldsymbol{\kappa} = \{\kappa_1, \dots, \kappa_L\}$, where $\kappa_l = (r_{attn}^{(l)}, r_{mlp}^{(l)})$. Biasing the initial population $P_0$ toward high-importance layers anchors the search in regions that preserve the representational capacity of the base model while adhering to the parameter budget $C$. This initialization strategy reduces search variance typically caused by stochastic sparsity assignments and prevents early convergence to architectures that fail to inherit pretrained capabilities.

To quantify the benefits of importance-based initialization, we conduct an ablation study on LLaMA-2-7B comparing our approach to uniform and random initialization strategies. Additionally, we compare our evolutionary search against random search in Table~\ref{tab:llama2_results_init}. Importance-based initialization provides a significantly stronger starting point than standard schemes; when initialized using $I_l$, \method{} achieves an average accuracy of 62.81\%, representing a 1.71\% improvement over uniform and a 2.18\% improvement over random initialization. These results indicate that leveraging layer-wise importance effectively warm-starts the search, accelerating the discovery of high-performing sub-networks under a fixed parameter budget.

Furthermore, evolutionary search yields a substantial 4.54\% absolute gain in average accuracy over random search. This improvement confirms that selecting elite candidates via $\Phi$, coupled with importance-based initialization, provides a robust search scheme for iterative refinement. By effectively navigating the discrete architectural space, this approach enables the consistent discovery of high performing models, without the high variance of random search. 

\subsection{Search Space Analysis}
\begin{figure*}[t]
    \centering
    \includegraphics[width=0.95\textwidth]{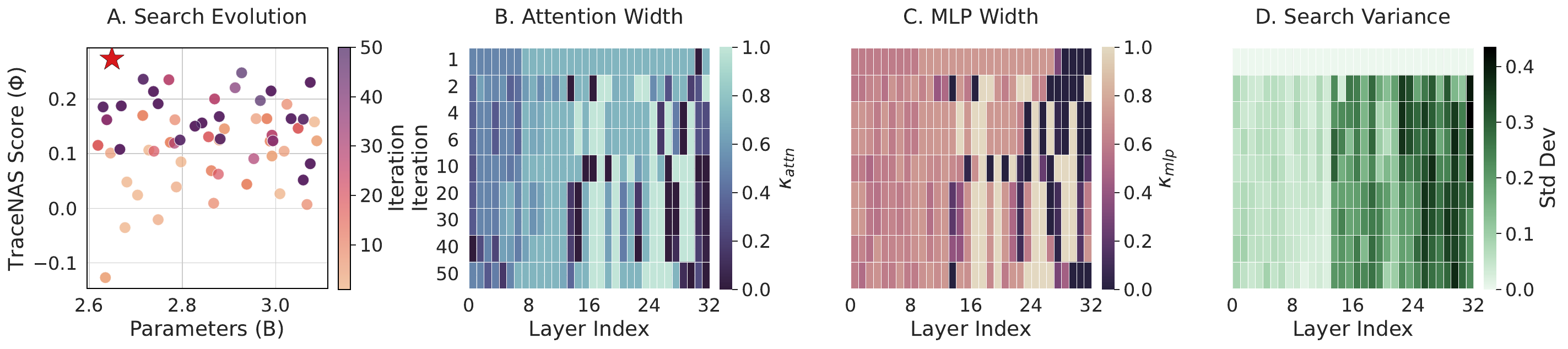}
    \caption{\textbf{\method{} Evolutionary Dynamics and Search Convergence.}
    Search trajectory for pruning Llama-2-7B to 2.7B across 50 iterations.
    (a) \method{} score $(\Phi)$ evolution: Illustrates the discovery of top scored candidates across the specified parameter budget window. The red star indicates the model with maximal functional inheritance for the given constraint.
    (b) Attention Width Evolution: Tracks the sparsity ratios for the attention sub-block $(\kappa_{attn})$; the search identifies specific layers where attention heads are critical for maintaining representational flow.
    (c) MLP Width Evolution: Sparsity ratios for MLP sub-blocks $(\kappa_{mlp})$, revealing the high structural sparsity and exploration across the model depth.
    (d) Total Search Variance: Illustrates the search variance across all identified models, the relative stability indicates that \method{} identifies high performing models within a short search window.  In figures (b), (c) and (d), $\kappa = 0.0$ indicates no pruning in that layer and $\kappa = 1$ indicates the layer has been dropped. }
    \vspace{-1ex}
    \label{fig:method_search_analysis}
\end{figure*}

To analyze the dynamics of the evolutionary search, Fig.~\ref{fig:method_search_analysis} illustrates the behavior of \method{} across 50 iterations. As shown in Fig.~\ref{fig:method_search_analysis}(a), the population moves from initial stochastic exploration toward candidates with progressively higher gradient alignment scores. The presence of negative alignment values in early iterations suggests that importance-based initialization alone cannot guarantee functional inheritance, as initial sparsity patterns introduce representational instability. This behavior also confirms that the prior does not overly constrain the search, leaving sufficient flexibility for the evolutionary process to explore the broad search space.

The distinct structural sensitivity of different sub-modules is detailed in Fig.~\ref{fig:method_search_analysis}(b) and (c). While the global importance prior is relatively uniform, the search discovers unique sparsity profiles for attention $(\kappa_{attn})$ and MLP $(\kappa_{mlp})$ widths. Specifically, the first iteration of the search finds that the final MLP layers are very important and thus they are not pruned, whereas attention modules less importance in later layers and thus initial layers have high sparsity. This indicates that representational integrity relies on different depth-wise configurations for different sub-module types.

Finally, Fig.~\ref{fig:method_search_analysis}(d) reports the population-level variance. The zero variance in the first iteration confirms that the search begins from a single anchored model, with diversity expanding only as mutations and crossovers are introduced in subsequent steps. The subsequent stabilization and variance plateau signal that the search successfully converged to a set of high-quality architectures. This trend validates the efficiency of the evolutionary mechanism in navigating the discrete search space and identifying optimal non-uniform configurations within a limited iteration budget.

\subsection{Evolution Search and CPT Hyperparameters}
\label{appendix:hyperparams}
Table~\ref{tab:wide_hyperparams} provides comprehensive overview of the evolution search and CPT hyperparameters used to realize the models presented in \method.
\begin{table*}[h]
    \centering
    \caption{Hyperparameter Configuration for \method{} Search (Left) and CPT (Right).}
    \label{tab:wide_hyperparams}
    \footnotesize
    \setlength{\tabcolsep}{5pt} 
    \begin{tabular}{lccccclccc}
        \toprule
        & \multicolumn{3}{c}{Search Parameters} & & & \multicolumn{3}{c}{CPT Parameters} \\
        \cmidrule(lr){2-4} \cmidrule(lr){7-9}
        \multirow{-2}{*}{Parameter} & 2.7B & 4.6B & 8.4B & & \multirow{-2}{*}{Parameter} & 2.7B & 4.6B & 8.4B \\
        \midrule
        Population Size ($P$) & \multicolumn{3}{c}{30} & & Learning Rate & \multicolumn{3}{c}{$1 \times 10^{-4}$} \\
        Elites ($K$) & \multicolumn{3}{c}{10} & & Batch Size & \multicolumn{3}{c}{1024} \\
        Crossover Rate & \multicolumn{3}{c}{0.7} & & LR Scheduler & \multicolumn{3}{c}{WSD} \\
        Mutation Rate &  \multicolumn{3}{c}{0.2 / 0.2} & & WSD Ratios & \multicolumn{3}{c}{0.05 / 0.65 / 0.30} \\
        Search Iterations ($N$) & \multicolumn{3}{c}{50} & & Context Length & 4,096 & 8,192 & 4,096 \\
        -- & \multicolumn{3}{c}{--} &  & Overall Tokens & \multicolumn{3}{c}{20B} \\
        \bottomrule
    \end{tabular}
\end{table*}

\subsection{Pseudocode}
\label{appendix: pseudocode}
Algorithm~\ref{alg:tracenas} provides detailed overview of the \method evolution search.
\begin{algorithm}[h]
\caption{\method}
\label{alg:tracenas}
\SetKwInOut{Input}{Input}
\SetKwInOut{Output}{Output}
\setstretch{0.9}

\Input{Pretrained $\mathcal{M}_{base}$; Budget $C$; Calibration set $\mathcal{B}$; Sparsity $\boldsymbol{\kappa}$; Elite size $k$; Search iterations $T$}
\Output{Optimal pruned sub-network $\hat{\mathcal{M}}_{sub}$}

\BlankLine
Initialize $\mathcal{M}_{base}$ with LoRA modules;\newline
$g_{base} \leftarrow \mathbb{E}_{b \in \mathcal{B}} [\nabla_{\theta} \mathcal{L}(\mathcal{M}_{base}(b; \theta))]$ \tcp*{Functional Anchor}
$I_l \leftarrow \mathbb{E}\big[ |W_l| \odot \|X_l\|_2 \big]$ \tcp*{Compute block importance for all $L$}

\BlankLine
Initialize $P_0$ with random depth $\mathbf{d} \in \{0,1\}^L$ and width $\boldsymbol{\kappa}$ weighted by $I_l$\;

\BlankLine
\For{$t = 1$ \KwTo $T$}{
    \For{each candidate $(\mathcal{M}_{sub}(\mathbf{d}, \boldsymbol{\kappa})) \in P_t$}{
        \tcp{Structural Realization}
        \uIf{$\#\text{Params}(\mathcal{M}_{sub}) > C$}{
            $\theta(\mathcal{M}_{sub}) \leftarrow -\infty$ \tcp*{Budget Penalty}
            \textbf{continue}\;
        }
        \Else{
            Realize masks $W'_{l} = W_{l} \odot \text{Mask}(\kappa_l)$ based on active layers $\mathbf{d}$\newline
            Route activations via residual connections for deactivated layers ($d_l=0$)\;
        }
        
        \BlankLine
        \tcp{Gradient-based Evaluation}
        $g_{sub} \leftarrow \mathbb{E}_{b \in \mathcal{B}} [\nabla_{\theta} \mathcal{L}(\mathcal{M}_{sub}(b; \theta))]$\;
        
        \For{each active layer $l$}{
            $\rho^{(l)} \leftarrow \text{Pearson Correlation Coefficient}(g_{sub}^{(l)}, g_{base}^{(l)})$ \tcp*{Gradient Trace Correlation}
        }
        
        \BlankLine
        \tcp{Fitness Aggregation}
        $\Phi(\mathcal{M}_{sub}) \leftarrow \sum_{l \in \text{Attn}} r_{attn}^{(l)} \rho^{(l)} + \sum_{l \in \text{MLP}} r_{mlp}^{(l)} \rho^{(l)}$\;
    }
    
    \BlankLine
    \tcp{Elite Selection and Reproduction}
    $E_t \leftarrow$ Select top $k$ candidates from $P_t$ based on $\Phi$\;
    $P_{t+1} \leftarrow E_t$ \tcp*{Carry over elite set directly}
    
    \While{$|P_{t+1}| < |P_t|$}{
        Parent $A, B \leftarrow$ Sample from $E_t$\;
        Child $C \leftarrow \text{Crossover}(A, B) + \text{Mutation}(C)$\;
        Add Child $C$ to $P_{t+1}$\;
    }
}
\Return $\hat{\mathcal{M}}_{sub} \leftarrow \mathop{\arg\max}\limits_{\Phi} \{P_T\}$
\end{algorithm}

\section{Motivation \& Conceptual Framework}
\subsection{Comparison with Influence Functions}
\label{appendix:compare_influence_functions}
To further clarify the positioning of \method, we distinguish our framework from data-centric influence functions~\cite{koh2017understanding,kwon2023datainf}, specifically LayerIF~\cite{askari2025layerif} and TracIn~\cite{pruthi2020estimating}. 

\subsection{Influence Functions}
Data influence functions (IF) are typically used to quantify how a specific training sample contributes to a model's prediction on a test sample. Specifically, IF quantifies how model parameters $\theta^*$ change when a specific training point $z_k$ is infinitesimally up-weighted. Formally, assuming the loss $\mathcal{L}$ is twice differentiable, the influence of the $k$-th training point on the parameters is defined as:
{\setlength{\abovedisplayskip}{4pt}
\setlength{\belowdisplayskip}{4pt}
$$I_{\theta^*} (z_k) := \frac{d\theta^{(k)}}{d\epsilon} \bigg|_{\epsilon=0} = -H(\theta^*)^{-1}\nabla_{\theta}\mathcal{L}(z_k, \theta)$$}
where $H(\theta)$ is the Hessian matrix. The influence of a training sample $z_i$ on the validation loss across $m$ validation points is then:
{\setlength{\abovedisplayskip}{4pt}
\setlength{\belowdisplayskip}{4pt}
$$I(z_i) = -\sum_{j=1}^{m} \nabla_{\theta}\mathcal{L}(z^V_j, \theta)^\top H(\theta)^{-1}\nabla_{\theta}\mathcal{L}(z_i, \theta)$$}

This measures whether a specific sample has a beneficial or detrimental impact on predictive performance.

\subsection{Structural Sensitivity via Data-Centric Influence}
LayerIF adapts this concept to assess the relative quality of different layers by localizing the computation to each layer $l$:
{\setlength{\abovedisplayskip}{4pt}
\setlength{\belowdisplayskip}{4pt}
$$I^{(l)}(z_i) = -\sum_{j=1}^{m} \nabla_{\theta}^{(l)} \mathcal{L}(z^V_j, \theta)^\top \left[H^{(l)}(\theta)\right]^{-1} \nabla_{\theta}^{(l)} \mathcal{L}(z_i, \theta)$$}

By aggregating positive influence scores, LayerIF derives a vector $S$ where each element $S^{(l)} = \sum_{i=1}^{n} \mathbb{I}[I^{(l)}(z_i) > 0] \cdot I^{(l)}(z_i)$, where $\mathbb{I}[\cdot]$ is the indicator function, captures the cumulative contribution of training data to validation performance through that specific layer. Crucially, LayerIF maintains a static model architecture; its analysis is data-centric, keeping the architecture fixed to evaluate layer specialization across different data samples.

\subsection{TracIn \& the Link to \method}
In contrast, \method{} keeps the data static and instead evaluates model sensitivity. We analyze the functional performance of the model as its architecture is structurally perturbed. Our formulation is inspired by TracIn, which simplifies standard influence functions by removing the $O(d^3)$ complexity of the inverse Hessian. TracIn computes a first-order approximation via the dot-product alignment between gradients of training and test samples:

{\setlength{\abovedisplayskip}{4pt}
\setlength{\belowdisplayskip}{4pt}
$$\text{TracIn}(z_i, z_j) = \sum_{t=1}^T \eta_t \nabla_{\theta} \mathcal{L}(z_i, \theta_t) \cdot \nabla_{\theta} \mathcal{L}(z_j, \theta_t)$$}

We extend this simplification to the domain of structural pruning by comparing the base model's gradient ($g_{base}$) and the pruned candidate's gradient ($g_{sub}$) on the same fixed data.

As detailed in Sec.~\ref{subsec:tracenas_score}, to quantify how effectively $\mathcal{M}_{sub}$ inherits the functional state of $\mathcal{M}_{base}$, we calculate the sub-block-wise Pearson Correlation Coefficient $\rho^{(l)}$:

{\setlength{\abovedisplayskip}{4pt}
\setlength{\belowdisplayskip}{4pt}
$$\rho^{(l)} = \frac{1}{N_l} \left\langle \left( \frac{g_{sub}^{(l)} - \mu_{g_{sub}^{(l)}}}{\sigma_{g_{sub}^{(l)}}} \right), \left( \frac{g_{base}^{(l)} - \mu_{g_{base}^{(l)}}}{\sigma_{g_{base}^{(l)}}} \right) \right\rangle$$}
where $N_l$ is the dimensionality of the low-rank subspace, $g_{base}$ and $g_{sub}$ are defined as $g = \mathbb{E}_{b \in \mathcal{B}} [\nabla_{\theta} \mathcal{L}(\mathcal{M}(b; \theta))]$, $\mu$ and $\sigma$ denote mean and standard deviation of gradient elements. By standardizing the traces $g$, $\rho^{(l)}$ captures the directional alignment decoupled from the magnitude shifts induced by pruning.

We aggregate these correlations using Sparsity-Weighted Aggregation to account for heterogeneous representation capacity:
{\setlength{\abovedisplayskip}{4pt}
\setlength{\belowdisplayskip}{4pt}
$$\Phi(\mathcal{M}_{sub}) = \sum_{l \in \text{Attn}} r_{attn}^{(l)} \cdot \rho^{(l)} + \sum_{l \in \text{MLP}} r_{mlp}^{(l)} \cdot \rho^{(l)}$$}

where $r^{(l)}$ is the retention ratio. This formulation anchors the global score in high-capacity regions, preventing $\Phi$ from being skewed by high-variance noise in aggressively pruned sub-blocks. This ensures the candidate resides within the original pretrained convergence basin, facilitating high-fidelity recovery.

\subsection{Why this works}
The primary reason this formulation works is that it treats the gradient as a local ``topographic map'' of the optimization landscape. While magnitude-based proxies like GradNorm measure if the landscape is stable, $\Phi$ measures if the gradient direction of the pruned model still points in the same direction as the original.

\begin{itemize}
    \item \textbf{Gradient Trace as a Proxy for the Optimization Path:} The gradient trace represents the direction of steepest descent. By calculating the alignment between $g_{base}$ and $g_{sub}$, we are measuring whether the sub-network wants to move toward the same local minima as the pretrained model. This is why we refer to it as Functional Inheritance, whether the pruned model inherits the optimization intent of its base model.
    \item \textbf{Decoupling Magnitude from Direction:} We used the Pearson Correlation Coefficient ($\rho^{(l)}$) specifically to decouple the directional signal from magnitude shifts. Structural pruning inherently reduces the total weight volume, which naturally suppresses gradient magnitudes. If we used a simple dot product, the score would drop simply because the model is smaller. Standardizing the traces via $\mu$ and $\sigma$ allows us to evaluate if the logic of the layer remains, regardless of its reduced power.
\end{itemize}

\section{Validation \& Proxy Correlation}
\label{appendix:downstream_correlation_analysis}
This section provides a detailed analysis of the \method{} proxy $\Phi$ and its ranking correlation with model perplexity (PPL), MMLU, and average downstream accuracies compared against the established baselines in Table~\ref{tab:proxy_correlation}.

\subsection{Correlation with Downstream Performance} 
As shown in Table \ref{tab:proxy_correlation}, \method{} achieves superior ranking correlation with downstream performance by effectively modeling functional inheritance. Proxies like NASWOT~\cite{mellor2021neural} and ZiCo~\cite{li2023zico} are formulated to  rank models based on expressivity and convergence capability from random initializations. NASWOT measures the linear separability of data representations by quantifying the dissimilarity in feature patterns using the Hamming distance. This metric characterizes the richness of data representations in a network, this is reflected in ranking performance on PPL $(\tau = 0.72)$. However, this does not account for whether a pruned architecture retains the specific linguistic knowledge, MMLU correlation of $(\tau=0.07)$ that is already encoded in the weights.

Similarly, ZiCo serves as a measure of loss landscape smoothness using the inverse coefficient of variation. This is defined as the ratio of the mean of gradients to their standard deviation across samples. This assumes that networks with high convergence speed and generalization capacity exhibit high absolute mean gradients and low variance, and will generalize well. This measure of generalizability is reflected in PPL $(\tau=0.5)$ ranking. These properties of ZiCo are crucial for optimization, yet they are inherently blind to the pretrained knowledge of extensively pretrained LLMs and the functional disruption caused by pruning, MMLU $\tau=0.26$,. These metrics optimize for a model that could learn well, whereas pruning requires a model that has already learned and preserved its pretrained distributions. This lack of accounting for pretrained knowledge inheritance explains their limited ranking correlation on specialized tasks like MMLU, where representational fidelity, not just trainability, is the primary driver of performance.

In contrast, GradNorm~\cite{abdelfattah2021zero} functions as a robust proxy for pruned variants of pretrained models. This stems from the principle it operates on: that the stability of a model, expressed through its gradient $L_2$ norm is a good measure of potential performance. Similarly, Synaptic Saliency~\cite{tanaka2020pruning} approximates parameter importance by measuring the impact of parameter removal on the model's loss function. Specifically, the proxy measures how much structural perturbations, in the form of parameter removal, impacts the model's total loss. However, these metrics primarily prioritize structural health and do not account for the global impact of pruning on representational depth or the specific disruption caused by multi-parameter pruning to knowledge retention tasks like MMLU. This limitation is reflected in their Kendall $\tau$ correlation values of only $\tau=0.45$ and $\tau=0.37$ on MMLU, respectively.

To further highlight the need for proxies to account for the global impact of pruning, we analyze MeCo~\cite{jiang2023meco}. MeCo measures generalization capacity of a model using the minium eigen values of the Pearson Correlation matrix across feature representations. This effectively captures generalizability, as reflected in PPL ranking ($\tau=0.79$). However, it does not translate to the complex reasoning and knowledge retention abilities required in pruned models, MMLU $\tau=0.02$, that inherit a distorted version of the pretrained weight state. Conversely, PrunerZero~\cite{dong2024pruner}, achieves the highest ranking correlation on MMLU, $\tau=0.69$. However, this high performance does not translate to model generalizability in the form of PPL or average accuracy on downstream tasks. We attribute this to the proxy formulation, defined as the product of the weight norms and the min-max scaled gradient vector. The high ranking correlation on MMLU is driven by the high magnitude weights acting as pointers to the high knowledge retention regions within the pretrained model. These regions are isolated by the min-max scaled gradients, making the proxy susceptible to outliers, thus resulting in poor performance on PPL $(\tau=0.31)$. Lastly, \#Params is a strong indicator of model generalizability through PPL ranking $(\tau=0.72)$. However, it does not account for representational collapse in pruned models, MMLU correlation $\tau=0.07$. Furthermore, it would not be able to distinguish between models under a contrained parameter budget. 

In contrast, \method{} provides the best end-to-end ranking correlation. While GradNorm is sensitive to mean directional shifts in gradients, \method{} employs centering via Pearson Correlation to isolate the functional gradient trace alignment from directional noise. This de-noises the signal to reveal the underlying structural inheritance that simpler magnitude-based metrics miss and functions as a reliable proxy for model performance. 

\subsection{Validating Sparsity-Weighted Pearson Correlation} 
To justify the formulation of $\Phi$, we evaluate dot product (\method{} - Dot), cosine similarity (\method{} - Cosine), and unweighted aggregation of Pearson correlation coefficients (\method{} - Unweighted) within the our low-rank gradient setup. We see that cosine similarity is more robust than dot product, which fails due to the magnitude shifts caused by structural pruning. However, cosine similarity remains sensitive to mean-gradient bias. Pearson correlation coefficient addresses this by centering the gradient traces, however, uniformly aggregating these coefficients across layers introduces instability in highly compressed blocks. By using layer sparsity as a weighting factor, $\Phi$ acts as a dynamic noise-filtering mechanism, thus de-emphasizing highly pruned layers. This weighting anchors the global score in the high-capacity regions identified as the primary repositories of functional inheritance. By prioritizing sub-blocks with higher parameter density, we prevent the global fitness signal from being skewed by the high-variance typical of aggressively pruned, low-capacity blocks. Furthermore, dot product fails as it lacks the scale-invariance necessary to handle the significant magnitude shifts inherently caused by structural pruning.

\subsection{Robustness of Proxy Stability: Correlation with Downstream Accuracy}
\label{appendix:corr_amoaballm_stability}

As shown in Sec.~\ref{subsec:proxy_stability}, the \method{} proxy $(\Phi)$ demonstrates strong robustness and inter hyperparameter stability in ranking pruned model performance across accuracy and perplexity. To further validate these results we show that $\Phi$, consistently shows high correlation with downstream performance metrics under various search constraints. We measure Spearman $\rho$ and Kendall $\tau$ correlations between the proxies (generated across all hyperparameters) and WikiText-2 Perplexity (PPL), MMLU accuracy and average accuracy in Table~\ref{tab:intra_proxy_correlation_rho} and~\ref{tab:intra_proxy_correlation_tau}.

\begin{table}[h]
    \centering
    \caption{Spearman $\rho$ correlation between various search hyperparameters and model PPL and average downstream accuracy. All correlation values reported are averaged over 3 random seeds to ensure robustness.}
    \label{tab:intra_proxy_correlation_rho}
    \footnotesize
    \begin{tabular}{l cc @{\hspace{2em}} l cc @{\hspace{2em}} l cc}
        \toprule
        \multicolumn{3}{c}{(a) $N$} & \multicolumn{3}{c}{(b) $CL$} & \multicolumn{3}{c}{(c) $r$} \\
        \cmidrule(r{2em}){1-3} \cmidrule(r{2em}){4-6} \cmidrule{7-9}
        $N$ & PPL & Avg Acc & $CL$ & PPL & Avg Acc & $r$ & PPL & Avg Acc \\
        \midrule
        4   & 0.893 & 0.832 & 128  & 0.932 & 0.864 & 2    & 0.923 & 0.900 \\
        8   & 0.925 & 0.883 & 256  & 0.915 & 0.834 & 8    & 0.933 & 0.902 \\
        16  & 0.930 & 0.902 & 512  & 0.896 & 0.820 & 64   & 0.930 & 0.902 \\
        64  & 0.923 & 0.896 & 1024 & 0.850 & 0.780 & 256  & 0.928 & 0.902 \\
        128 & 0.928 & 0.903 & 2048 & 0.915 & 0.864 & 1024 & 0.928 & 0.902 \\
        256 & 0.841 & 0.748 & 4096 & 0.930 & 0.902 & 4096 & 0.955 & 0.912 \\
        \bottomrule
    \end{tabular}
\end{table}

\begin{table}[h]
    \centering
    \caption{Kendall $\tau$ correlation between various search hyperparameters and model PPL and average downstream accuracy. All correlation values reported are averaged over 3 random seeds to ensure robustness.}
    \label{tab:intra_proxy_correlation_tau}
    \footnotesize
    \begin{tabular}{l cc @{\hspace{2em}} l cc @{\hspace{2em}} l cc}
        \toprule
        \multicolumn{3}{c}{(a) $N$} & \multicolumn{3}{c}{(b) $CL$} & \multicolumn{3}{c}{(c) $r$} \\
        \cmidrule(r{2em}){1-3} \cmidrule(r{2em}){4-6} \cmidrule{7-9}
        $N$ & PPL & Avg Acc & $CL$ & PPL & Avg Acc & $r$ & PPL & Avg Acc \\
        \midrule
        4   & 0.742 & 0.680 & 128  & 0.803 & 0.708 & 2    & 0.791 & 0.745 \\
        8   & 0.790 & 0.732 & 256  & 0.779 & 0.689 & 8    & 0.803 & 0.748 \\
        16  & 0.797 & 0.748 & 512  & 0.750 & 0.671 & 64   & 0.797 & 0.748 \\
        64  & 0.778 & 0.736 & 1024 & 0.712 & 0.636 & 256  & 0.795 & 0.746 \\
        128 & 0.790 & 0.739 & 2048 & 0.767 & 0.706 & 1024 & 0.794 & 0.748 \\
        256 & 0.707 & 0.613 & 4096 & 0.797 & 0.748 & 4096 & 0.837 & 0.760 \\
        \bottomrule
    \end{tabular}
\end{table}
The results confirm that \method{} maintains a stable and predictive ranking signal across a wide range of hyperparameter settings, validating that the inherited optimization landscape of the base model is reliably captured without requiring dense calibration. We provide a detailed analysis of the results in Table~\ref{tab:intra_proxy_correlation_rho}:
\begin{itemize}
    \item Calibration Sample Density ($N$): With as few as 4 calibration samples, the proxy achieves a strong correlation of 0.832 with average accuracy. This correlation peaks and stabilizes between $N=16$ and $N=128$ at approximately 0.90, supporting the use of relatively low sample counts during evolutionary search to maximize efficiency without sacrificing ranking fidelity. Interestingly, the correlation dips noticeably at N=256, which may result from additional samples introducing less optimal data points that could dilute the alignment proxy. This suggests a potential benefit from filtering or weighting calibration samples to maintain high-quality proxy scores.
    \item Context Length ($CL$): Proxy correlations remain high at shorter context windows; for example, at $CL=128$, the PPL correlation is 0.932 and the accuracy correlation is 0.864. Increasing context length to $CL=4096$ further improves alignment, reaching an accuracy correlation of 0.902. This indicates that while local dependencies are captured early, longer contexts enhance the proxy’s ability to predict complex reasoning performance.
    \item Low-Rank Gradient Subspace ($r$): The proxy shows exceptional stability across gradient subspace ranks. Even at a very low rank of $r=2$, correlation with average accuracy remains at 0.900, validating the Intrinsic Dimensionality Invariance principle: the core manifold dynamics necessary for effective pruning exist within a compact subspace. Although correlation improves slightly to 0.912 at full rank ($r=4096$), the minimal gain does not justify the additional overhead, confirming the effectiveness of using lower-rank subspaces during search.
\end{itemize}
By maintaining approximately 0.90 correlation with downstream accuracy across these hyperparameters, \method{} provides a scalable and reliable foundation for zero-shot model compression. This stability ensures that the identified sub-networks consistently reside within the original pretrained convergence basin regardless of search-time resource limitations.

\subsection{Performance Across Different Sparsities}
\label{appendix:ppl_analysis}
\begin{wrapfigure}{r}{0.3\textwidth}
\centering
\vspace{-3ex}
\includegraphics[width=0.3\textwidth]{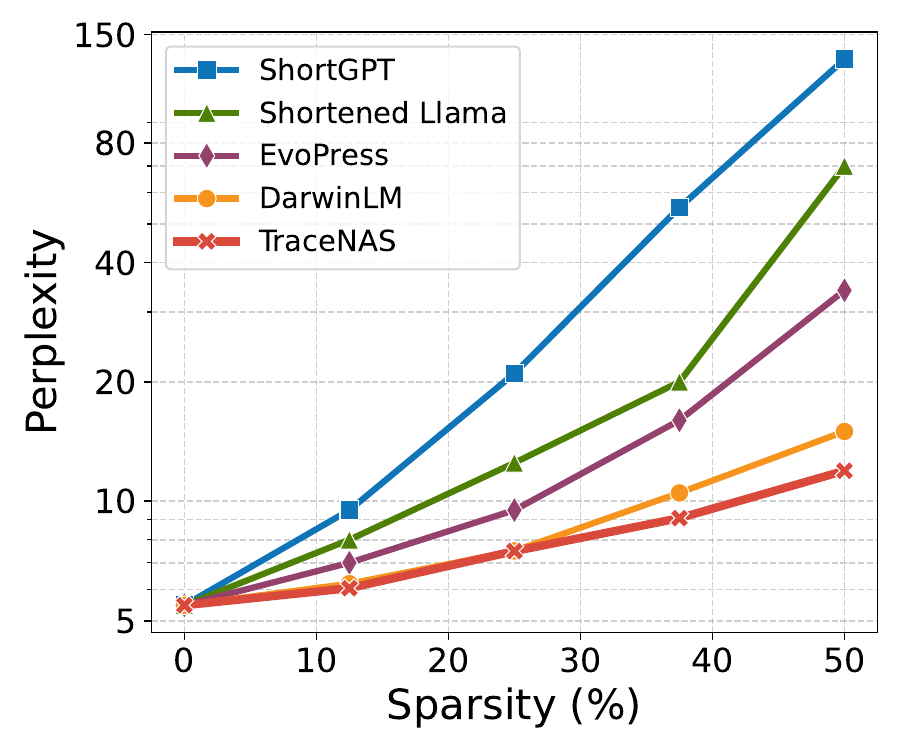}
\caption{\textbf{\method{} PPL across sparsity levels.} WikiText-2 perplexity reported for pruned models identified via \method{} evolutionary search and trained using 2.5B tokens of CPT.}
\vspace{-5ex}
\label{fig:method_sparsity_ppl}
\end{wrapfigure}
We evaluate \method{} across a range of sparsity levels and report WikiText-2 perplexity for LLaMA-2-7B in Fig.~\ref{fig:method_sparsity_ppl} after lightweight CPT with 2.5B tokens. We generate compressed models ranging between 7B (0\% sparsity) and 3.3B (50\% sparsity) parameters using \method{} over 200 evolutionary search iterations on 16 sequences of data from FineWeb-Edu. Across all sparsity regimes, \method{} consistently achieves lower perplexity than competing width and layer pruning baselines.

As expected, perplexity increases with sparsity for all methods; however, \method{} exhibits a flatter degradation curve. At 50\% sparsity, \method{} attains a perplexity of approximately 11, compared to roughly 15 for DarwinLM and over 100 for ShortGPT. This resilience under aggressive compression suggests that the gradient trace proxy effectively identifies architectures that remain within the pretrained model’s convergence basin, mitigating representational collapse as parameters are removed.

Notably, the robustness of \method{} at high sparsity arises from architectural selection rather than additional recovery training. All methods are evaluated under comparable lightweight CPT budgets, yet \method{} consistently identifies sub-networks with substantially lower perplexity by prioritizing architectures that preserve optimization landscape. \method{} leverages gradient alignment to capture global sensitivity across layers, enabling the retention of long-range dependencies even under severe compression. The key advantage of \method{} lies in its ability to discover non-uniform architecture configurations tailored to a given parameter budget. Whereas layer-dropping approaches primarily remove entire blocks in a uniform manner, \method{} jointly optimizes depth and per-layer width, retaining parameters where they are most necessary. By guiding the search using gradient alignment the resulting sub-networks maintain geometric alignment with the base model’s optimization landscape, yielding significantly more stable performance as sparsity increases.

\section{Scalability \& Speedup Analysis}
\subsection{Speedup Analysis}
\label{appendix:speedup_analysis}
\begin{table}[ht]
\centering
\caption{Inference phase specific speedup analysis on NVIDIA A100. $S_p$ and $S_d$ denote the prefill and decode speedup, respectively.}
\label{tab:master_results}
\footnotesize
    \begin{tabular}{lcccccccc}
    \toprule
    Model & TTFT & TPOT & Throughput & Peak Memory & $S_{p}$ & $S_{d}$ \\
     & (ms) & (ms) & (T/s) & (GB) & (Prefill) & (Decode) \\
    \midrule
    Llama-2 7B & 1462.05 & 35.80 & 27.93 & 15.01 & 1.00$\times$ & 1.00$\times$ \\
    ShearedLlama 2.7B & 3071.05 & 33.24 & 47.92 & 7.88 & 0.48$\times$ & 1.08$\times$ \\
    DarwinLM 2.7B & 649.83 & \textbf{12.96} & \textbf{83.50} & \textbf{6.07} & 2.25$\times$ & \textbf{2.76}$\times$ \\
    \method{} 2.7B (Ours) & 346.56 & 17.61 & 56.80 & 6.33 & \textbf{4.22}$\times$ & 2.03$\times$ \\
    \midrule
    Llama-3.1 8B & 960.25 & 42.55 & 23.50 & 16.52 & 1.00$\times$ & 1.00$\times$ \\
    Minitron 4.4B & 1340.86 & 34.61 & 31.70 & 10.49 & 0.72$\times$ & 1.23$\times$ \\ 
    DarwinLM 4.6B & 918.68 & \textbf{18.95} & \textbf{56.54} & 11.53 & 1.05$\times$ & \textbf{2.25}$\times$  \\ 
    \method{} 4.6B (Ours) & 556.52 & 26.61 & 37.58 & \textbf{9.98} & \textbf{1.73}$\times$ & 1.60$\times$ \\ 
    \midrule
    Qwen 2.5 14B & 1927.08 & 79.43 & 12.59 & 30.29 & 1.00$\times$ & 1.00$\times$ \\
    Minitron 8.4B & 5361.37 & 72.19 & 20.05 & 18.03 & 0.36$\times$ & 1.10$\times$ \\ 
    DarwinLM 8.4B & 4717.75 & \textbf{46.18} & \textbf{29.99} & 19.88 & 0.41$\times$ & \textbf{1.72}$\times$ \\ 
    \method{} 8.4B (Ours) & 1322.30 & 54.52 & 18.34 & \textbf{17.71} & \textbf{1.46}$\times$ & 1.46$\times$ \\
    \bottomrule
    \end{tabular}
\end{table}
The practical advantages of the \method{} architectures are detailed in Table~\ref{tab:master_results}. We evaluate inference performance on an NVIDIA A100 GPU using a 4096 token prompt and a 128 token generation window. To isolate architectural effects under a uniform execution path, all models are evaluated using native PyTorch with a standardized inference configuration. Results are recorded over 20 independent trials to reduce runtime variance and ensure stable, comparable measurements. As shown in the results, our architectures demonstrate significant improvements in memory efficiency and interactive latency, primarily driven by reductions in TTFT.

While training-aware baselines like DarwinLM demonstrate higher raw decode throughput ($S_d$), our models achieve a substantial reduction in Time To First Token (TTFT), which is a critical metric for interactive responsiveness. Specifically, at the 2.7B scale, \method{} reaches a 4.22$\times$ prefill speedup ($S_p$), representing a 1.97$\times$ improvement over DarwinLM’s prefill speedup. This trend of superior prefill efficiency persists across larger scales, suggesting that our gradient alignment objective effectively identifies layer configurations that minimize the computational overhead of the initial prompt processing phase. 

Furthermore, \method{} architectures exhibit superior memory efficiency, maintaining the lowest peak memory utilization at the 8.4B and 14B parameter scales. For instance, our 8.4B configuration requires only 17.71 GB of peak memory, outperforming both Minitron and DarwinLM. While \method{} models do not maximize raw decode throughput, they prioritize TTFT and memory efficiency, which are more critical for real-time, user-facing deployment.

\subsection{Extending \method{} to Different Model Scales}
\label{appendix:tracenas_model_scales}
\begin{table*}[h]
\centering
\caption{Pruning results for Llama-3.1-70B Instruct.}
\label{tab:llama3.1_70b_results}
\vspace{-1ex}
\begin{footnotesize}
\addtolength{\tabcolsep}{-3.5pt} 
\begin{tabular}{l l c ccccc ccc c}
\toprule
\multirow{-1}{*}{Method} & \multirow{-1}{*}{\#Params} & MMLU  & PIQA & WG & ArcE & ArcC & HS & SciQ & LQA & BQ & \multirow{-1}{*}{Avg.} \\
\midrule
Llama-3.1-70B Instruct & 70B & 81.22 & 83.89 & 85.9 & 87.2 & 70.39 & 87.44 & 96.12 & 37.21 & 85.2 &  79.17 \\
\midrule
PUZZLE & 51B & \textbf{80.20} & -- & 84.53 & -- & \textbf{69.20} & \textbf{85.58} & -- & -- & -- & --  \\
DarwinLM & 35B & -- & 81.2 & 83.5 & 82.5 & 60.5 & 80.3 & 95.4 & 33.0 & 84.2 & 75.07 \\
\method{} (Ours) & 40B & 77.21 &\textbf{82.1} & \textbf{85.21} & \textbf{83.19} & 60.32 & 78.21 & \textbf{97.32} & \textbf{34.12} & \textbf{85.12} & \textbf{75.70} \\
\bottomrule
\end{tabular}
\end{footnotesize}
\end{table*}

\begin{table*}[h]
\centering
\caption{Pruning results for Pythia-2.8B and Gemma-2-2B .}
\label{tab:pythia_gemma_results}
\vspace{-1ex}
\begin{footnotesize}
\addtolength{\tabcolsep}{-3.5pt} 
\begin{tabular}{l l c ccccc ccc c}
\toprule
\multirow{-1}{*}{Method} & \multirow{-1}{*}{\#Params} & MMLU  & PIQA & WG & ArcE & ArcC & HS & SciQ & LQA & BQ & \multirow{-1}{*}{Avg.} \\
\midrule
Pythia-2.8B  & 2.8B & 24.30 & 74.04 & 60.37 & 64.43 & 35.65 & 60.73 & 88.4 & 28.41 & 63.48 & 59.48 \\
\midrule
DarwinLM & 1.4B & 25.1 & 71.3 & 57.3 & 61.2 & 34.7 & 54.5 & 82.9 & 27.9 & 65.0 & 56.85 \\
\method{} (Ours) & 1.4B & \textbf{25.5} & \textbf{71.8} & \textbf{58.69} & \textbf{62.83} & 34.4 & 54.11 & \textbf{84.9} & \textbf{28.18} & \textbf{64.92} & \textbf{57.47} \\
\midrule
\midrule
Gemma2-2B  & 2.5B & 51.35 & 79.27 & 71.66 & 80.21 & 53.32 & 74.65 & 95.7 & 29.18 & 72.72 & 69.58 \\
\midrule
DarwinLM & 1.2B & 25.3 & 61.3 & 52.1 & 48.5 & 23.2 & 30.5 & 80.0 & 26.4 & 55.5 & 47.18 \\
\method{} (Ours) & 1.2B & \textbf{26.44} & \textbf{70.34} & \textbf{57.93} & \textbf{66.70} & \textbf{36.77} & \textbf{51.84} & \textbf{91.2} & \textbf{27.49} & \textbf{65.62} & \textbf{58.23} \\
\bottomrule
\end{tabular}
\end{footnotesize}
\end{table*}
We demonstrate that \method{} consistently identifies high-performing sub-networks across a wide range of model scales in Tables~\ref{tab:llama3.1_70b_results} and~\ref{tab:pythia_gemma_results}. At the 70B scale, pruning Llama-3.1-70B to 40B parameters, \method{} maintains strong reasoning performance, with MMLU accuracy of 77.21 and a SciQ accuracy of 97.32. These results show that \method{} gradient alignment proxy effectively navigates the large-scale architecture space without search-time training. These are results are achieved after performing post-pruning supervised finetuning (SFT) for 5000 steps using LoRA~\cite{hu2022lora} and the Orca~\cite{mukherjee2023orca} finetuning dataset, further showcasing that \method{} can be used as a proxy for finetuning performance. 

The benefits of \method{} are equally evident in smaller models. For Pythia-2.8B pruned to 1.4B parameters, \method{} improves average performance over DarwinLM (57.47 vs. 56.85) and achieves the highest accuracy on nearly all reasoning tasks. For Gemma2-2B pruned to 1.2B parameters, \method{} dramatically reduces the performance drop seen in DarwinLM, achieving an average of 58.23 compared to 47.18 while leading across nearly all individual tasks. The results reported for the 2B models are after for CPT on FineWeb-Edu 100BT for 10B tokens. 

Overall, these results highlight that \method{} is robust across scales. By leveraging gradient alignment to guide architectural selection, it consistently produces non-uniform sub-networks that outperform training-aware pruning approaches, for both massive 70B models or highly compressed 1.2B models.

\subsection{Training with more tokens}
\label{appendix:token_scales}
We provide results on training our Llama-2-2.7B pruned model on 10B tokens, 20B tokens and 50B tokens and compare with baselines ShearedLlama and DarwinLM. 
\begin{table*}[h]
\centering
\begin{threeparttable}
\caption{\method{} LLaMA-2-2.7B model trained across 10B, 20B and 50B tokens on FineWeb-Edu 100BT subset.}
\label{tab:llama2_results_token_scales}
\vspace{-1ex}
\small
\renewcommand{\arraystretch}{1.0} 
\setlength{\tabcolsep}{5pt} 
\begin{tabular}{l c cc | ccccc ccc c}
\toprule
Method & \#Params & Srch & Recv & PIQA & WG & ArcE & ArcC & HS & SciQ & LQA & BQ & Avg. \\
\midrule
Llama-7B & 6.7B & -- & -- & 77.69 & 74.11 & 76.34 & 52.81 & 78.96 & 93.80 & 29.80 & 77.73 & 70.16 \\
\midrule
ShearedLLaMA & 2.7B & 0.4B & 50B & \textbf{75.80} & 64.20 & 67.00 & 41.20 & \textbf{70.80} & 90.80 & 28.20 & 63.00 & 62.63 \\
DarwinLM & 2.7B & 1.0B & 10B & 73.12 & 63.77 & 69.94 & 43.94 & 66.95 & 90.20 & 28.57 & \textbf{64.10} & 62.57 \\
\midrule
\method{} (Ours) & 2.7B & 98M & 10B & 71.54 & 59.43 & 68.25 & 40.24 & 66.40 & 89.40 & 28.57 & 62.01 & 60.73 \\
\method{} (Ours) & 2.7B & 98M & 20B & 73.40 & 63.97 & 70.20 & 43.80 & 68.54 & 91.10 & \textbf{29.71} & 61.76 & 62.81 \\
\method{} (Ours) & 2.7B & 98M & 50B & 73.95 & \textbf{64.35} & \textbf{70.93} & \textbf{44.67} & 69.12 & \textbf{91.28} & 28.18 & 62.42 & \textbf{63.11} \\
\bottomrule
\end{tabular}
\end{threeparttable}
\end{table*}

\subsection{Evaluating \method{} for Unstructured Pruning}
\label{appendix:compare_unstructured}
We evaluate the capability of \method{} as a proxy for unstructured pruning of the dense Llama-2-7B model and evaluate its performance after SFT. We compare 50\% pruned \method{} model against magnitude pruning~\cite{han2015deep}, FLAP~\cite{an2024fluctuation}, Wanda~\cite{wang2020structured}, ShortenedLlama~\cite{kim2024shortened}, AmoebaLLM~\cite{fu2024amoeballm} and PrunerZero~\cite{dong2024pruner} in Table~\ref{tab:llama2_unstructured}.

We finetune our 50\% pruned model on the Orca~\cite{mukherjee2023orca} dataset for 10,000 steps with LoRA~\cite{hu2022lora}, AmoebaLLM SFT on Alpaca for 10,000 steps, 50\% pruned and finetuned Wanda model, 50\% pruned FLAP model and the reported ShortenedLlama results. Evaluating \method{} under this setup showcases its robustness as a proxy for performance potential under unstructured pruning and SFT post-pruning recovery.  
~\begin{table*}[h]
\centering
\caption{Evaluating Llama-2-7B unstructured pruning using \method{} against SOTA unstructured pruning baselines. Average accuracy excludes MMLU.}
\label{tab:llama2_unstructured}
\vspace{-1ex}
\begin{footnotesize}
\addtolength{\tabcolsep}{-3.5pt} 
\begin{tabular}{l l c cccc c c}
\toprule
Method & \#Params & MMLU & WG & ArcE & ArcC & HS & BQ & Avg. \\
\midrule
Llama-2-7B & 6.7B & 77.69 & 76.34 & 52.81 & 78.96 & 93.80 & 70.16 & 74.41 \\
\midrule
ShortenedLlama & 50\% & 24.76 & 51.54 & 50.63 & 26.45 & 43.60  &  62.23 & 46.89 \\
Magnitude & 50\% & 30.69 & 59.19 & 58.84 & 33.53 & 45.49 & 54.59 & 50.33 \\
FLAP & 50\% & 27.67 &  56.75 & 55.43 & 31.57 & 51.33  & 59.45  & 50.91 \\
AmoebaLLM & 50\% & 30.60 &  60.10 & 56.60 & 31.50 & 51.30  &  65.70 & 53.04 \\
Wanda & 50\% & \textbf{39.27} & 68.19 & 72.77 & 39.59 & 52.49 & \textbf{75.99} & 61.81 \\
PrunerZero & 50\% & -- & 67.80 & 69.19 & 38.82 & \textbf{69.12} & 68.87  & 62.76 \\
\method{} (Ours) & 50\% & 38.76 & \textbf{68.83} & \textbf{73.86} & \textbf{42.51} & 56.42 & 74.13 & \textbf{63.15} \\
\bottomrule
\end{tabular}
\end{footnotesize}
\vskip -0.1in
\end{table*}

The results in Table~\ref{tab:llama2_unstructured} showcase that \method{} has the highest average accuracy at 63.15\% outperforming Wanda (61.81\%) and PrunerZero (62.76\%). The high performance of Wanda showcases why we chose it as a mask generation metric and our higher performance further validates that \method{} significantly boosts performance over Wanda. 


\section{Generation Quality Example}
\label{appendix: gen_example}

\begin{table}[H]
\centering
\small
\renewcommand{\arraystretch}{1.2} 
\caption{\textbf{Qualitative Analysis:} TraceNAS (2.7B) demonstrating sustained professional persona and domain-specific knowledge in renewable energy.}
\label{tab:qualitative_results}
\begin{tabular}{|p{0.95\columnwidth}|} 
\hline
\textbf{Prompt:} Act as a Renewable Energy Consultant specializing in grid modernization. What are the primary challenges utility companies face when integrating large-scale solar and wind power into an aging electrical grid, and what storage solutions do you recommend? \\ \hline
\method: 
The main challenge is that there’s no one solution that can meet the needs of all customers. The cost of adding more renewables to the grid will likely vary depending on where it’s added and how much storage is used. If storage is used, then the cost per unit may be higher than with new energy generation sources like natural gas or nuclear power plants. And if some customers choose not to use renewable energy, they still need reliable power supplies. \\
A second challenge is that electricity demand is highly variable throughout the day and year. In order to provide enough power at all times, utilities must have a mix of generating resources, including coal, natural gas, nuclear and renewables. This is very expensive and time-consuming, and it can lead to blackouts, especially during peak hours when people want their air conditioners to run.
What is your favorite renewable energy project to date? \\
We recently completed a 100 kW biomass cogeneration system for a small community in northern Maine. Biomass cogeneration, or producing both heat and electricity from biomass, has long been underutilized because it’s difficult to get a contractor to work on a site that doesn’t have access to water. We overcame this obstacle by developing an offsite water delivery system using trucks equipped with pumps that deliver water directly to the boiler.
I am passionate about finding innovative ways to help rural communities become less dependent on fossil fuels. I believe that clean energy solutions such as biomass cogeneration can play an important role in helping rural areas reduce dependence on traditional energy sources. \\ \hline
\end{tabular}
\end{table}

\section{Limitations}
\method{} provides an efficient zero-shot metric for model pruning, yet several constraints remain. The proxy $\Phi$ captures immediate functional alignment but does not explicitly model loss landscape curvature or smoothness, factors which may influence long-term convergence. Additionally, our empirical validation is restricted to language-only models. Although the framework is extensible, its performance on multi-modal architectures has not yet been verified. From a deployment perspective, our search process does not incorporate hardware-specific latency bottlenecks or inference-engine optimizations such as vLLM. Furthermore, we do not perform a full multi-objective search to map the entire Pareto frontier between size, speed, and accuracy, focusing instead on validating the $\Phi$ proxy under fixed architectural constraints. Future work will explore hardware-aware metrics and cross-modal generalizability to broaden the utility of zero-shot structured pruning.




\end{document}